\documentclass{elsarticle}
\usepackage[total={6in, 10in}]{geometry}

\usepackage[utf8]{inputenc} 
\usepackage[T1]{fontenc}
\usepackage[hidelinks]{hyperref}    
\usepackage{lineno}
\modulolinenumbers[2]
\usepackage{amsmath}
\usepackage{amssymb}
\usepackage{mathtools}
\usepackage{acro}
\usepackage{booktabs}
\usepackage[table]{xcolor}
\usepackage{collcell}
\usepackage{hhline}
\usepackage{makecell}
\usepackage{multirow}
\usepackage{pgf}
\usepackage{rotating}
\usepackage{tikz}
\usepackage{color}
\usepackage{todonotes}

\usepackage{localmacros}

\def\colorModel{hsb} 
\newcommand{\minvalue}{30.5} 
\newcommand{\maxvalue}{35.0} 

\newcommand\ColCell[1]{
  \pgfmathparse{(min(max(#1,\minvalue),\maxvalue)-\minvalue)/(\maxvalue-\minvalue)}
  \edef\colorval{\pgfmathresult} 
  
  \pgfmathparse{#1<(\minvalue+\maxvalue)/2?1:0}  

  \ifnum\pgfmathresult=0\relax\color{white}\fi
  \ifdim\colorval pt < 0.7 pt
    \edef\brightness{0}
   \else
      \edef\brightness{\colorval/4}
  \fi
  \pgfmathsetmacro\compA{0} 
  \pgfmathsetmacro\compB{\colorval} 
  \pgfmathsetmacro\compC{1-\brightness} 
  \edef\x{\noexpand\cellcolor[\colorModel]{\compA,\compB,\compC}}\x #1
}
\newcolumntype{E}{>{\collectcell\ColCell}c<{\endcollectcell}}
\newcolumntype{H}{>{\centering\arraybackslash}p{0.8cm}} 

\DeclareAcronym{LPD}{
  short = LPD,
  long = learned primal-dual}
\DeclareAcronym{iLPD}{
  short = iLPD,
  long = invertible learned primal-dual}
\DeclareAcronym{LPDh}{
 short = LPDh,
 long = learned primal-dual for helical geometry}
\DeclareAcronym{LSPD}{
  short = LSPD,
  long = learned stochastic primal-dual}
\DeclareAcronym{CT}{
  short = CT,
  long = computed tomography}
\DeclareAcronym{DNN}{
  short = DNN,
  long = deep neural network}
\DeclareAcronym{CNN}{
  short = CNN,
  long = convolutional neural network}
\DeclareAcronym{PSNR}{
  short = PSNR,
  long = peak signal-to-noise ratio}
\DeclareAcronym{SSIM}{
   short = SSIM,
   long = structural similarity index}
   \DeclareAcronym{SD}{
   short = SD,
   long = source-detector}
\DeclareAcronym{2D}{
   short = 2D,
   long = two-dimensional}
\DeclareAcronym{3D}{
   short = 3D,
   long = three-dimensional}
\DeclareAcronym{FBP}{
   short = FBP,
   long = filtered back-projection}
\DeclareAcronym{PDHG}{
   short = PDHG,
   long = primal-dual hybrid gradient}
\DeclareAcronym{ODL}{
   short = ODL,
   long = Operator Discretization Library}
\DeclareAcronym{GT}{
   short = GT,
   long = ground truth}

\begin{document}

\begin{frontmatter}



\title{Sparse View Tomographic Reconstruction of Elongated Objects using Learned Primal-Dual Networks}

\author[uns]{Buda Baji\'{c}}
\ead{buda.bajic@uns.ac.rs}

\author[ltu]{Johannes A. J. Huber \corref{cor1}}
\ead{johannes.huber@ltu.se}
\cortext[cor1]{Corresponding author}

\author[ltu]{Benedikt Neyses}
\ead{benedikt.neyses@ltu.se}

\author[ltu]{Linus Olofsson}
\ead{linus.olofsson@ltu.se}

\author[kth]{Ozan Öktem}
\ead{ozan@kth.se}

\affiliation[uns]{
    organization={Faculty of Technical Sciences, University of Novi Sad},
    addressline={Trg Dositeja Obradovića 6},
    city={Novi Sad},
    postcode={21000},
    country={Serbia}}

\affiliation[ltu]{
    organization={Wood Science and Engineering, Luleå University of Technology},
    addressline={Forskargatan 1}, 
    city={Skellefteå},
    postcode={93177}, 
    country={Sweden}}

\affiliation[kth]{
    organization={Department of Mathematics, KTH Royal Institute of Technology},
    addressline={Lindstedtsvägen 25}, 
    city={Stockholm},
    postcode={10044},
    country={Sweden}}

\begin{abstract}
In the wood industry, logs are commonly quality screened by discrete X-ray scans on a moving conveyor belt from a few source positions. Typically, the measurements are obtained in a single \ac{2D} plane (a "slice") by a sequential scanning geometry. The data from each slice alone does not carry sufficient information for a three-dimensional tomographic reconstruction in which biological features of interest in the log are well preserved. In the present work, we propose a learned iterative reconstruction method based on the Learned Primal-Dual neural network, suited for sequential scanning geometries. Our method accumulates information between neighbouring slices, instead of only accounting for single slices during reconstruction.
Evaluations were performed by training U-Nets on segmentation of knots (branches), which are crucial features in wood processing.
Our quantitative and qualitative evaluations show that with as few as five source positions our method yields reconstructions of logs that are sufficiently accurate to identify biological features like knots (branches), heartwood and sapwood. 
\end{abstract}


\end{frontmatter}

\section{Introduction}

Many industrial applications of tomography involve scanning objects that move on a conveyor belt \cite{de2014industrial}. 
A specific setting in this context is the sequential scanning geometry where the acquired tomographic data of the moving object relates to a specific \ac{2D} cross-section. 
A common \ac{2D} \ac{CT} set up consists of multiple fixed \ac{SD} pairs, or of a single \ac{SD} pair that rotates around the object. 
In order to minimize equipment costs and to make fast acquisition applicable to real-time industrial needs, industrial \ac{CT} scanners typically contain either a limited number of fixed \ac{SD} pairs or a single rotating \ac{SD} pair which takes a small number of source positions for X-ray emission. 
This results in only few measurements per \ac{2D} slice (sparse view data). 
Such data may not be sufficient to reconstruct a high-quality tomographic \ac{2D} image. Furthermore, if one aims to obtain a full \ac{3D} reconstruction, then information between slices needs to be combined to provide sufficient angular information of the scanned object.

The wood processing industry presents a case where sparse view reconstruction may be valuable. Detailed knowledge of the interior volume of a wooden log and its quality-determining biological features can increase the yield and the value of the products extracted from a log. Today, this is achieved by full X-ray \ac{CT} scanners using a helical scanning geometry with hundreds or even thousands of measurements per rotation, enabling high-speed (2-3~m/s) scanning at sufficient resolution for detecting relevant features in logs. The first scanner has been in operation in a Swedish sawmill since 2017. 
More common in sawmills, especially smaller ones, are however discrete X-ray scans of logs from a few static angles along a conveyor belt, which is attractive due to significantly lower investment cost compared to a \ac{CT} scanner.
It can be assumed that nearby \ac{2D} cross-sections (slices) of the log along its length axis will be similar because the interior structure of the log only varies slowly in that direction.
This circumstance may enable the use of sparse view reconstruction methods that make use of the similarity among neighbouring slices.

This study explores an idea of using information among sparsely sampled \ac{2D} slices in order to obtain a tomographic image from only few source positions per slice, which is sufficiently accurate to distinguish the quality-determining biological features, such as knots (branches), heartwood and sapwood. Heartwood is the biologically inactive and relatively dry central part of the trunk, and sapwood is the actively water and nutrient transporting part, in which the distinction between knot tissue and surrounding wood can be severely obscured due to the high moisture content.
This idea is inspired by dynamic process analysis where certain phenomena evolve in time \cite{hakkarainen2019undersampled}.
Here, in a sequential scanning of a moving object, we assume analogously that \ac{2D} slices are evolving along the direction of the movement, i.e. the conveyor belt (a third dimension). 
More specifically, the goal of this study is to extend the \ac{2D} \ac{LPD} method (see section \ref{sec:related_work}) for tomographic reconstruction by accumulating the information along the third dimension of the object based on the assumption that neighbouring \ac{2D} cross-sections of the object are similar. 

We aim to propose a method that can reconstruct 3D volume information similar to that obtained with a full CT scanner, but with 2 -- 3 orders of magnitude fewer measurements. The approach should be applicable to elongated objects with only gradual changes of their features along the length axis. In this study we evaluate our method for the case of pine wood logs, with pine being one of the most important wood species for industrially produced wood products.

This paper is organized as follows. In section \ref{sec:related_work} we discuss the tomographic reconstruction task and give an overview of different model and data-driven approaches which address it. Section \ref{sec:method} gives the theoretical foundation for deep learning-based reconstruction methods. Further, it presents the original \ac{LPD} method and its extension adapted for sequential scanning in an industrial setting proposed in this paper. Next, we present the experimental setup in section \ref{sec:evaluation} and results for the wood industry in section \ref{sec:results}. Finally, we conclude in section \ref{sec:conclusion}.

\section{Related work}\label{sec:related_work}

\subsubsection{Sparse-view reconstruction}

Tomography refers to a wide range of methods for imaging the interior of an object by probing it from different directions with a penetrating wave or particle.
The image that one seeks to recover corresponds in this context to the interior \ac{2D} or \ac{3D} structure of the object being studied. Tomographic image reconstruction is then the task of computationally recovering the image from noisy indirect observations (observed tomographic data).

Image reconstruction in X-ray \ac{CT} has traditionally relied on the \ac{FBP} method and variants thereof. These are analytical methods which compute a regularised approximate inverse of the ray transform. Unfortunately, the \ac{FBP} assumes dense angular sampling and does not provide useful reconstruction results under sparse scanning geometries. 
In addition, it is problem-specific and does not generalize well to different acquisition geometries encoded in the forward/back-projection model.

A natural way to overcome this drawback was to formulate a general class of reconstruction methods that allow to replace the forward operator in a plug-and-play manner. This leads to variational models, where the reconstruction task is formulated as a minimization problem of some cost function that ensures data-consistency and incorporates a regularizer which gives robust reconstructions with desirable properties. The regularizer incorporates \textit{a priori} knowledge about the solution, commonly different sparsity assumptions \cite{rudin1992nonlinear,donoho2006compressed,daubechies2004iterative}. Handcrafting a sufficiently descriptive regularizer and ensuring computational feasibility (due to the large-scale numerical computations) are the main challenges related to this iterative model-based method. In addition, selecting an optimal regularization parameter which balances data-consistency and the regularizer is also of critical importance for the performance of the variational-based model.

The development of data-driven reconstruction methods is motivated by the need to address the above challenges. Data-driven methods are learning an optimal reconstruction from the training data instead of handcrafting it. \emph{Learned reconstruction} is defined as a mapping between the data and image space and it is parametrized by some suitably chosen \ac{DNN}. The key challenge here is how to choose a \ac{DNN} in the best way.
One natural idea is to select a generic \ac{DNN} architecture, consisting of fully connected layers. However, when such architectures are applied to \ac{CT} data, they tend to become very large due to the input and output sizes. Another approach is to apply a \ac{DNN} architecture on a hand-crafted approximate inverse of the forward operator (e.g. \ac{FBP}) as an image-to-image \emph{post-processing} operator that is learned from training data. Popular architectures in imaging for this are based on \acp{CNN}, like U-Net \cite{jin2017deep}. 

A different way to domain adaptation is achieved through \emph{unrolling} \cite{monga2021algorithm}. 
The approach starts with an iterative scheme, such as one designed to minimize data-discrepancy in variational models. 
The next step is to truncate this scheme and replace the handcrafted updates with (possibly shallow) \acp{CNN} (unrolling). 
The reconstruction method is then a \ac{DNN} that is formed by stacking these (shallow) \acp{CNN} and accompanying them with physics-driven operators, 
and a forward operator and its adjoint, which are explicitly given (non-learned) \cite[Sec.~4.9.1]{arridge2019solving}. 
An example of this is the \ac{LPD} method \cite{adler2018learned} with its variants \cite{rudzusika2021invertible,rudzusika20223d,tang2021stochastic} and variational networks \cite{hammernik2018learning}. 

Another possibility for learned reconstruction is to deploy a generative deep learning framework such as diffusion models \cite{webber2024diffusion}, which have recently gained popularity.

All data-driven methods are suited exclusively to a \ac{2D} or \ac{3D} scanning geometry, while none of them specifically targets the industrial sequential \ac{2D} scanning geometry, to the best of our knowledge. In this study, we aim to extend the state-of-the-art \ac{2D} learned iterative \ac{CT} image reconstruction \ac{LPD} method to be applicable to a sequential \ac{2D} scanning geometry. This will be achieved by accumulating the information along the third dimension of the object assuming that neighbouring \ac{2D} cross-sections of the object are similar. A related approach can be found in \cite{springer2022reconstruction} where a dimension-reduced Kalman Filter \cite{hakkarainen2019undersampled} is used to accumulate information between consecutive slices, which has been applied in wood log scanning \cite{Gergel:2019aa}. The main difference between the method presented in \cite{springer2022reconstruction} and the 2.5D~\ac{LPD} proposed here is that the former is model-driven, while the latter is data-driven.

\subsubsection{CT-based methods in the wood industry}

\ac{CT} scanning enables precise measurements of the internal density field in logs, and industrial CT systems are now routinely employed for log sorting and to optimise log orientation prior to sawing \cite{Rais2017,ursella2018}.
Numerous studies have demonstrated that CT data is well suited for image analysis techniques to detect key anatomical features, including the pith location and knots  \cite{LONGUETAUD2012,JOHANSSON2013}, pitch pockets, heartwood and sapwood, earlywood and latewood, as well as grain patterns \cite{Seplveda2002,Ekevad2004}. 
Similarly, the radial, tangential and longitudinal material directions in clearwood have been reconstructed from CT images using image analysis methods \cite{Hansson2016,Huber2022}.

Collectively, this body of research underscores the potential of CT data to yield detailed anatomical, structural and potentially material property information.
With growing computational capabilities, the 3D density field can be used for deriving numerical models of the scanned objects, e.g. to evaluate moisture relations \cite{Hansson2016} or estimate the mechanical performance using computational mechanics \cite{Huber2022}.

\section{Method}\label{sec:method}

First, we provide the mathematical formalization for deep learning-based reconstruction methods. Then we outline the original \ac{LPD} architecture. Lastly, we describe the proposed  2.5D~\ac{LPD} method, which extends \ac{LPD} method and is suited for a sequential scanning geometry.

\subsection{Mathematical formalization}

\begin{table}[htbp]
\centering
\caption{Summary of mathematical notation used}
\label{tab:notation}
\begin{tabular}{cl}
\toprule
\textbf{Symbol} & \textbf{Description} \\
\midrule
    $\Real$ & Set of real numbers \\[2pt]
    $\Natural$ & Set of natural numbers \\[2pt]
    $\RecSpace$ & Reconstruction (image) space \\[2pt]
    $\DataSpace$ & Data (measurement) space \\[2pt]
    $\DataManifold$ & Measurement manifold (lines for ray transform) \\[2pt]
    $\ForwardOp$ & Forward operator (ray transform) \\[2pt]
    $\ForwardOp^*$ & Adjoint of the forward operator \\[2pt]
    $\stimage$ & True image (random variable) \\[2pt]
    $\image^*$ & Ground-truth image (realization of $\stimage$) \\[2pt]
    $\image$ & Reconstruction (estimate of image, primal variable) \\[2pt]
    $\stdata$ & Observed data (random variable) \\[2pt]
    $\data$ & Measured tomographic data (realization of $\stdata$) \\[2pt]
    $\stdatanoise$ & Observation error/noise (random variable) \\[2pt]
    $\datanoise$ & Realization of observation noise \\[2pt]
    $\dual$ & Dual variable \\[2pt]
    $\param$ & Parameters of neural networks \\[2pt]
    $\hat{\param}$ & Optimal learned parameters \\[2pt]
    $\RecOp_{\param}$ & Learned reconstruction operator parameterized by $\param$ \\[2pt]
    $\Loss$ & Loss function \\[2pt]
    $\Expect$ & Expectation \\[2pt]
    $\PrimalMappingOriginal_{\param_k^p}$ & CNN-based primal update operator at iteration $k$ \\[2pt]
    $\DualMappingOriginal_{\param_k^d}$ & CNN-based dual update operator at iteration $k$ \\[2pt]
    $\NumUnrollingIterates$ & Number of unrolled primal-dual iterations \\[2pt]
    $\Dim$ & Number of preceding iterations used \\[2pt]
    $\numTData$ & Number of training data pairs \\[2pt]
    $\totcross$ & Number of consecutive cross-sections considered \\[2pt]
    $\text{PSNR}$ & Peak Signal-to-Noise Ratio \\[2pt]
    $\text{SSIM}$ & Structural Similarity Index \\[2pt]
\bottomrule
\end{tabular}
\end{table}

Tomographic data formation can be defined as 
\begin{equation}\label{eq:InvProbStatistical}
 \stdata = \ForwardOp(\stimage) + \stdatanoise, 
\end{equation}
where $\stimage$ is an $\RecSpace$-valued random variable that generates the true (unknown) image $\image^* \in \RecSpace$, the random variable $\stdatanoise$ models the observation error, and the observed noisy tomographic data $\data \in \DataSpace$ is a single sample of a $\DataSpace$-valued (conditional) random variable $(\stdata \mid \stimage = \image^*)$. Here, $\ForwardOp \colon \RecSpace \to \DataSpace$ is the forward operator (ray transform) that models how a signal generates data in absence of noise and observation errors.

The traditional aim in tomographic reconstructions is to recover the entire (posterior) distribution of the $\RecSpace$-valued random variable $(\stimage \mid \stdata = \data)$. Since this is too demanding in imaging applications, a natural variant is to recover an estimator that summarises the posterior, or alternatively, to sample from it. More formally, a \emph{reconstruction method} can be defined as a parametrized mapping $\RecOp_{\param} \colon \DataSpace \to \RecSpace$ which corresponds to a statistical estimator and has a desired property that 
\[ [ \ForwardOp \circ \RecOp_{\param} ](\data) \approx \data
   \quad\text{for data $\data \in \DataSpace$ and suitable choice of $\param>0$.}
\] \emph{Learned reconstructions} $\FullMethod$ are typically parameterized by some suitably chosen \ac{DNN}. Therefore, \textit{learning} here refers to selecting optimal parameters $\hat{\param}$ from the training data. In this case, we assume that we have access to pairs of ground-truth images and the corresponding data.  It means that training data $\mathcal{D}$ are given as i.i.d. samples $(\image^1, \data^1), \dots, (\image^\numTData, \data^\numTData) \in \RecSpace \times \DataSpace \; \text{of} \; (\stimage, \stdata)$.
\emph{Supervised training} is performed by minimizing the loss functional
\begin{equation} \label{eq:supervised_training}
    \text{L}(\param) = \Expect_{(\stimage,\stdata) \sim \mathcal{D}} \bigl[\Loss_{\RecSpace}(\FullMethod(\stdata), \stimage) \bigr]
\end{equation} 
for training dataset $\mathcal{D}$. 
Here $\Loss_{\RecSpace} \colon \RecSpace \times \RecSpace \to \Real$ measures how close the obtained reconstruction is to the ground truth, i.e., it quantifies consistency in image space $\RecSpace$. 
A typical choice for the loss function is the squared $\ell_2$-norm $\Loss_{\RecSpace}(\image, \image') :=  \Vert\image-\image' \Vert_2^2$, resulting in a reconstruction method which approximates the posterior mean. 
Another option can be to use $\ell_1$-loss $\Loss_{\RecSpace}(\image,\image'):= \Vert \image-\image' \Vert_1$ and this choice can similarly be interpreted as having a learned reconstruction method $\FullMethod(\data)$ approximating the posterior median.

\subsection{Original \ac{LPD} method}\label{sec:LPD_original}

First we introduce the state-of-the art learned iterative method for CT image reconstruction - Learned Primal-Dual (\acs{LPD}) \cite{adler2018learned}. The original \ac{LPD} method integrates elements of model- and data-driven approaches for solving ill-posed inverse problems. The combination of ideas from classical regularization theory and recent advances in deep learning enables to perform learning while making use of prior information about physical modeling of the inverse problem.

The Learned Primal-Dual architecture is a domain adapted neural network which is typically trained in a supervised manner (Equation~\eqref{eq:supervised_training}) with $\ell_2$ loss. The \ac{LPD} architecture is inspired by the iterative scheme in the \ac{PDHG} algorithm \cite{chambolle2011first}. This architecture incorporates a forward operator into a deep neural network by unrolling a proximal primal-dual optimization scheme and replacing proximal operators with \acp{CNN}. 
More precisely, the \ac{LPD} architecture is given in Algorithm~1, the number of unrolling iterates is $\NumUnrollingIterates$, $\ForwardOp^*$ is the adjoint of the forward-operator, while functions $\DualMappingOriginal_{\param^d_k}$ and $\PrimalMappingOriginal_{\param^p_k}$ correspond to \acp{CNN} with different learned parameters but with the same architecture for each unrolled iteration $k$. 
The primal and dual spaces are extended by letting $\Dim$ be greater than $1$, which allows the algorithm to also use preceding iterations, akin to having ``memory'', instead of just using the last iteration, i.e. $\image = [\image^{(0)}, \image^{(1)}, \dots, \image^{(C)}] \in \PrimalSpace^\PrimalDim$ and the same holds for the dual variable.
Superscripts (1) and (2) denote the first and second channels of the assigned variables. Therefore, the number of input channels of the primal $\PrimalMappingOriginal_{\param^p_k}$ and dual mapping $\DualMappingOriginal_{\param^d_k}$ \acp{CNN} are respectively $\Dim+1$ (inputs are the $\Dim$ previous estimates of the primal variable and the adjoint operator is applied on its first channel) and $\Dim+2$ (inputs are the $\Dim$ previous estimates of the dual variable and the forward operator is applied on its second channel and the data $\data$), and the number of output channels is $\Dim$ for both. For more details, see \cite{adler2018learned}.

\begin{table}[t]
\centering
	\small
	\begin{tabular}[t]{@{\;}l}
	\toprule
	\textbf{Algorithm~1} \Acs{LPD} \cite{adler2018learned} \\
	\hline
	1: Choose initial primal and dual variables \\
	$(\primal_0, \dual_0)=\texttt{init}(\data)$, where $(\primal_0, \dual_0) \in (\PrimalSpace^\PrimalDim, \DataSpace^\DualDim)$ \\
	2: \textbf{For} $k=1,2,\dots,\NumUnrollingIterates$ \textbf{do}: \\
    3: \hspace{0.3cm} Dual update: $\dual_{k} = \dual_{k-1} + \DualMappingOriginal_{\param^d_k} \left(\dual_{k-1}, \ForwardOp (\primal_{k-1}^{(2)}), y\right)$ \\
	4: \hspace{0.3cm} Primal update: $\primal_{k} = \primal_{k-1} +  \PrimalMappingOriginal_{\param^p_k} \left(\primal_{k-1}, \ForwardOp^\ast (\dual_k^{(1)})\right)$ \\
	5: \textbf{return} $x_M^{(1)}$ \\
    \hline
	\end{tabular}
\end{table}

In the original \ac{LPD} paper \cite{adler2018learned}, the method is successfully applied for \ac{CT} reconstruction with 2D geometries. It has recently been further scaled up \cite{rudzusika2021invertible,rudzusika20223d,tang2021stochastic} for application on 3D geometries in clinical \ac{CT} scanners.

\subsection{2.5D~\ac{LPD} reconstruction method}\label{sec:2.5D_LPD}

In this work we continue extending the original \ac{LPD} method with the motivation to make it applicable for industrial use, particularly for the case when \ac{2D} sequential scanning is used with only few \ac{SD} pairs. 
We first formulate the mathematical problem of reconstructing \ac{CT} images from data in this scenario and then introduce the new \ac{LPD} adaptation tailored for it.

\subsubsection{Sequential measurement setting}
The \ac{2D} reconstruction and data spaces are respectively
\[ \RecSpace = \{ \text{\ac{2D} images} \}
   \quad\text{and}\quad
   \DataSpace = \{ \text{\ac{2D} sinograms} \}.
\]
The aim is now to recover a series of \ac{2D} cross-sections $\image = \{\image_{i}\}_{i=1}^{\totcross} \in \RecSpace^\totcross$ from corresponding series of 2D tomographic data $\data = \{\data_{i}\}_{i=1}^{\totcross} \in \DataSpace^\totcross$ where  
\[
  \data_j = \ForwardOp_j(\image_j) + \datanoise_j
  \quad\text{for $j=1,\ldots, \totcross$.}
\] 
Here, $\totcross \in \Natural$ denotes the number of cross-sections/data pairs accounted for (in ``memory''), $\image_j \in \RecSpace$ is the $j$-th 2D cross-section, $\data_j \in \DataSpace$ is the corresponding 2D sparse view tomographic data with $\datanoise_j \in \DataSpace$ denoting the (unknown) observation error, and $\ForwardOp_j \colon \RecSpace \to \DataSpace$ is the forward operator associated with the data $\data_j$. 
The latter is essentially the \ac{2D} ray transform sampled on some known manifold $\DataManifold_j$ of lines that traverse the object and lie in the same \ac{2D} plane as the cross-section $\image_j$. 
Thus, $\ForwardOp_j$ is the ray transform restricted to lines in $\DataManifold_j$. 
Note also that the manifold $\DataManifold_j$ varies with $j=1,\ldots,\totcross$.

\subsubsection{2.5D~\ac{LPD} architecture}

The main idea for taking into account similarity between neighbouring cross-sections is to use the ``memory'' in the \ac{LPD} architecture described at the end of section \ref{sec:LPD_original}.
This allows us to update a single cross-section by using several neighbouring preceding cross-sections instead of using only the previous estimate of the certain single cross-section. 

More formally, we define the reconstruction operator $\RecOp_{\param} \colon \DataSpace^{\totcross} \to \RecSpace$ as $\RecOp_{\param}(\data) := \hat{\image}_{\totcross}$ where the 2D cross-section $\hat{\image}_{\totcross} \in \RecSpace$ is given by the following iterative scheme for both the primal and dual variables:
\begin{equation*}
\begin{cases}
   \{ \dual_{i} \}_{i=1}^{\totcross}
    :=  \{\dual_{i} \}_{i=1}^{\totcross} + \DualMappingOriginal_{\param^d_k}\Bigl(
        \{ \dual_{i} \}_{i=1}^{\totcross},
        \bigl\{ \ForwardOp_{i}(\image_{i}) \bigr\}_{i=1}^{\totcross},
        \{ \data_{i} \}_{i=1}^{\totcross}
     \Bigr)
\\[0.75em]
   \{ \image_{i} \}_{i=1}^{\totcross}
    := \{ \image_{i} \}_{i=1}^{\totcross} + \PrimalMappingOriginal_{\param^p_k}\Bigl(
        \{ \image_{i} \}_{i=1}^{\totcross},
        \bigl\{ \ForwardOp^{\ast}_{i}(\dual_{i}) \bigr\}_{i=1}^{\totcross}
     \Bigr)
\end{cases}
\end{equation*}
for $k=1, \ldots, \NumUnrollingIterates$ with $\param = (\param^d, \param^p)=(\param^d_1, \ldots, \param^d_{\NumUnrollingIterates}, \param^p_1,\ldots,\param^p_{\NumUnrollingIterates})$ and the primal and dual mappings
\begin{align*}
 \DualMappingOriginal_{\param^d_k} &\colon \DataSpace^{\totcross} \times \DataSpace^{\totcross} \times \DataSpace^{\totcross} \to \DataSpace^{\totcross} 
 \\
 \PrimalMappingOriginal_{\param^p_k} &\colon \RecSpace^{\totcross} \times \RecSpace^{\totcross} \to \RecSpace^{\totcross} 
\end{align*}
are \acp{CNN} that are not necessarily deep.
The primal and dual mapping \acp{CNN} have the same architecture but different learned parameters for each unrolled iteration $k$, which corresponds to the original \ac{LPD}. Note that the number of input channels of the primal and dual mapping \acp{CNN} are respectively $3\times \totcross$ and $2\times \totcross$, and the number of output channels is $\totcross$ for both.
For the implementation details, please refer to \ref{app:LPD_implementation_details}.

The above iterative scheme outputs a finite sequence $\{ \hat{\image}_{i} \}_{i=1}^{\totcross} \in \RecSpace^{\totcross}$ whose entries correspond to estimates of the 2D cross-sections at $1, 2, \ldots,\totcross$. Hence, in particular we obtain an estimate $\hat{\image}_{\totcross} \in \RecSpace$ of the last 2D cross-section in that sequence.

\subsubsection{Training protocol}

The training protocol is based on having access to supervised training data $\mathcal{D}$ of the form $(\image^j,\data^j) \in \RecSpace^\totcross \times \DataSpace^\totcross$ for $j=1,\ldots,\numTData$ where 
\[ \image^j = (\image_1^j, \ldots, \image_\totcross^j)
   \quad\text{and}\quad    
   \data^j = (\data_1^j, \ldots, \data_\totcross^j)
   \quad\text{with}\quad    
   \data_i^j \approx \ForwardOp_i(\image_i^j).
\]
Then, the trained 2.5D~\ac{LPD} $\RecOp_{\hat{\param}} \colon \DataSpace^{\totcross} \to \RecSpace$ can be given by solving the following learning problem where the hyper-parameter $\hat{\param}$ is obtained as 
\[
   \hat{\param} \in \argmin_{\param} 
   \frac{1}{\numTData}\sum_{j=1}^{\numTData} \bigl\Vert \RecOp_{\param}(\data^j) - \image_\totcross^j \bigr\Vert_2^2.
\]
Note that minimizing this loss function is the same as minimizing the loss functional given by Equation.~\eqref{eq:supervised_training} with $\ell_2$ loss and expectation being estimated by the arithmetic mean for the given data set $\mathcal{D}$.

\section{Evaluation}\label{sec:evaluation}

\subsection{Log dataset}
The Swedish \textit{stem bank} \cite{gronlund1995} is a comprehensive database comprising several hundred Scots Pine (\textit{Pinus sylvestris L.}) and Norway Spruce (\textit{Picea abies L. Karst.}) trees, established in 1995 at the Division of Wood Science and Engineering at Luleå University of Technology in Skellefteå in collaboration with AB Trätek. 
The trees were lengthwise subdivided into logs (between 4--5\ m long), resulting in 628 pine and 750 spruce logs, which were characterized by a variety of measurements.

At the time of the creation of the stem bank, each log was scanned using a Siemens SOMARIS-ART medical CT scanner using a 5~mm tick fan-beam and a proprietary algorithm (SP9) for reconstruction. 
Scanning was conducted sequentially on a slice-by-slice basis along the length of each log and to achieve economically viable scanning times, the creators of the stem bank decided to conduct scans at 10~mm intervals in the feature-rich and highly lengthwise varying regions near \textit{knot-whorls} (groups of knots), and at 40~mm intervals in the remaining regions of the logs where the lengthwise variability was low.
Before storing the acquired image stacks in the stem bank, the scans were resampled by linear interpolation to a consistent lengthwise resolution of 10~mm, with $256 \times 256$ pixel cross-sections and a bit depth of 8-bit.


From the stem bank, we chose at random pine logs to comprise the disjoint sets ``set A'' and ``set B'', where the latter have been manually annotated by humans for knot labels using the MONAI Label framework \cite{monai2024} (Figure~\ref{fig:dataset}). In radiographic images, the distinction between late knot tissue and water-rich sapwood is challenging for human eyes, and likewise the exact point of appearance of a knot root, which needs to be considered in our evaluations.

\subsection{Validation of the \ac{LPD} algorithm}

For the initial evaluation of the 2.5D~\ac{LPD} algorithm we used log cross-sections from set B (split 1 in Figure ~\ref{fig:dataset}), with $42$ log volumes used for training, one log volume for validation, and three for testing. Each log contained approximately $400$ slices.
From each slice, \ac{2D} data (i.e. the sinogram) was simulated applying a forward operator from the \ac{ODL} \cite{adler_odl} with a fan-beam geometry using $360$ source positions. 
The resulting data was further sparsified such that only few source positions were left. 

We experimented with the number of source positions to study how small their number could be made while still yielding a satisfactory reconstructed image. Answering this question is important from the perspective of fast acquisition by preferably cheap industrial \ac{CT} scanners with a limited number of \ac{SD} pairs. 
For the demonstration in this paper, we chose the number of source positions being equal to $5$, $7$, $9$, $11$, $13$ and $15$. For the 2.5D~\ac{LPD} method, neighbouring slices are obtained from different source positions. 
To consider the ideal case, in which two consecutive acquisition geometries have a random angular increment, which maximises information, we uniformly sampled the difference between consecutive sources $\Delta$ between $0$ and $360$ degrees.
In addition, this random choice reflects a likely scenario in a sawmill conveyor, where logs may roll and jump uncontrolled, and could potentially also rotate simultaneously.
Assume e.g. $5$ sources, then $\Delta=360/5=72$ which will result in the sources being located at $(0, 72, 144, 216, 288)$ degrees, and at a (randomly chosen) angular increment of $7$ degrees, these locations will be moved to $(7, 79, 151, 223, 295)$ degrees in the next slice. 

We evaluated the 2.5D~\ac{LPD} with the number of neighbouring cross-sections $\totcross$ varying from $2$ to $9$ in order to explore how the number of consecutive slices being taken into account affects the reconstruction quality.

In addition to the strategy where the last slice in a sequence of several consecutive slices is reconstructed by 2.5D~\ac{LPD}, we explored a ``middle'' strategy where the middle slice is reconstructed instead. The motivation for this strategy is to improve the reconstruction of the beginning of knots, which should benefit from the ability of the method to \textit{look ahead} a few slices, instead of only \textit{looking back}.

To quantitatively evaluate the validity of 2.5D~\ac{LPD}, we measured the closeness of the reconstruction with the original CT image in terms of \ac{PSNR} and \ac{SSIM} \cite{wang2004image}. The higher either of these measures are, the closer the reconstructed image corresponds to the original CT image.

\subsection{Evaluation by knot segmentation}\label{sec:knotsegmentation}

The quality of sawn timber is predominantly affected by knots and their positioning inside the sawn volume.
Inside and in the vicinity of knots, the fibre orientations of wood are severely deviating from its otherwise rather mildly varying orientation parallel to the growth direction of the tree, i.e. in \textit{clearwood}.
In the regions of deviating fibre orientations the mechanical properties are impaired in comparison to clearwood, e.g. the stiffness and strength of softwoods can be approximately 30-fold weaker in direction across the fibres as compared to along the fibres.

The industrial value of tomographic reconstructions from sparse data needs to be evaluated against its suitability to serve as a base for an automated detection of interesting biological features, and in particular knots.
We therefore evaluated the suitability of the \ac{LPD} based reconstructions for knot segmentation.
Since the purpose of knot segmentation in our context is solely to evaluate the quality of the \ac{LPD} reconstructions, we specifically chose a ubiquitous, off-the-shelf \ac{CNN} based architecture, since \acp{CNN} have previously been used effectively on CT images for this task \cite{ursella2018}.
Hence, we employed the MONAI U-Net \cite{monai2024}, an enhanced version of the classical U-Net, incorporating residual units and efficient dimension matching \cite{residualUnet2019}.
For its implementation details, please refer to \ref{app:U-Net_implementation_details}.

\begin{figure}[ht!]
\centering
\includegraphics[width=1\linewidth]{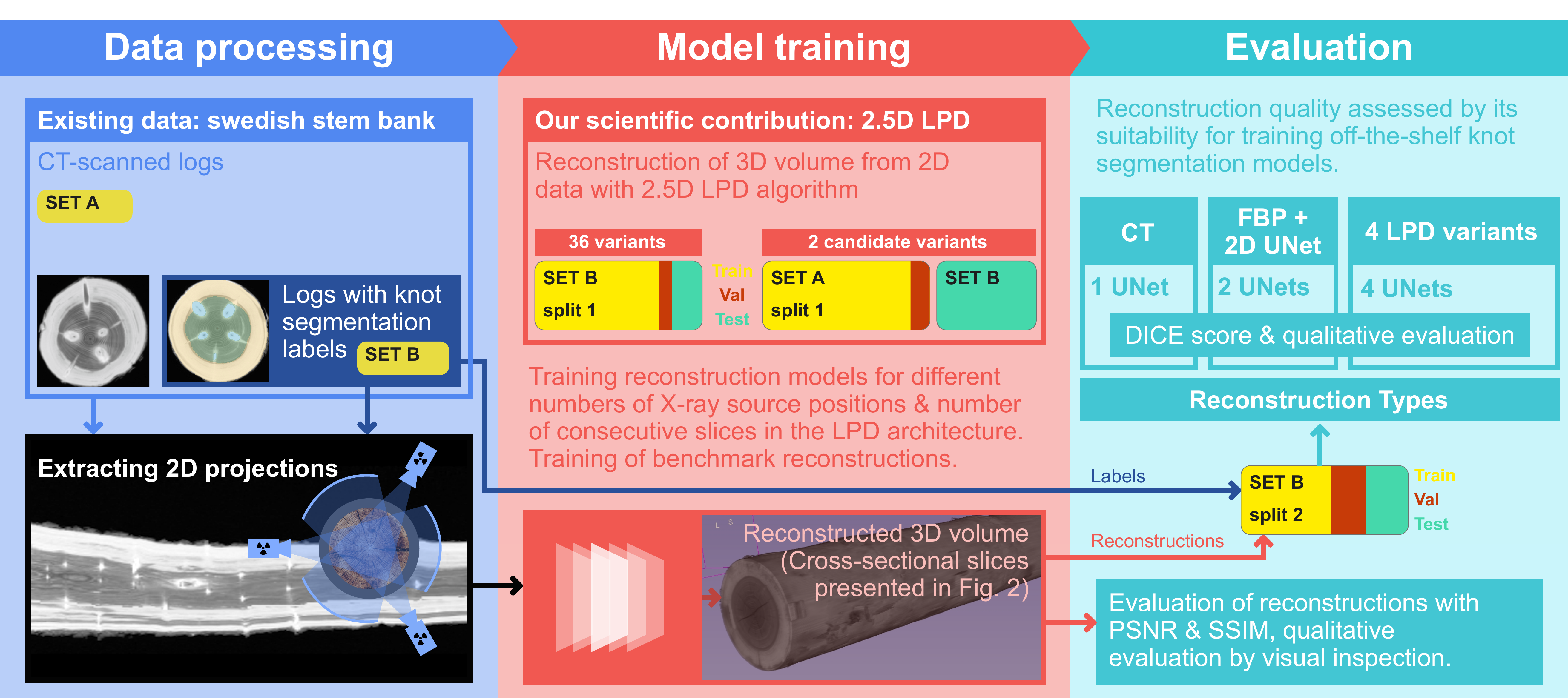}
\caption{Overview of the experimental workflow and dataset structure.}
\label{fig:dataset}
\end{figure}

Seven U-Nets for knot segmentation were trained (Figure~\ref{fig:dataset}): one based on the full \ac{CT} reconstruction, two on reference reconstruction methods employing 2D U-Net post-processing of \ac{FBP} reconstructions (FBP \& 2D UNet) as in \cite{jin2017deep}, and four based on what we deemed the most suitable candidates for \ac{LPD}-based reconstruction in an industrial setup based on results presented in Section~\ref{subsec:PSRN_res}.
The U-Net post-processed FBP reference variants and the \ac{LPD}-based candidates used either 5 source positions, in the plain 2D \ac{LPD} fashion (2D \ac{LPD} 5-pos), and additionally with 5 consecutive slices employing the ``middle'' slice 2.5D \ac{LPD} strategy (2.5D \ac{LPD} 5-pos 5-slices), or 9 source positions, with plain 2D (2D \ac{LPD} 9-pos), and additionally 3 consecutive slices employing the ``last'' slice strategy (2.5D \ac{LPD} 9-pos 3-slices).
Each of those U-Nets were trained on the respective reconstructions of each method for the same logs from set B (split 2 in Figure ~\ref{fig:dataset}), with $42$ samples used for training, $5$ for validation and $4$ for testing, where each sample represents a whole log.

The \ac{GT} knot labels were the same for all cases.
The FBP \& 2D UNet and the candidate 2D and 2.5D \ac{LPD} methods were specifically retrained from scratch for the segmentation evaluation using set A (split 1 in Figure ~\ref{fig:dataset}), with the cross-sections of $60$ logs for training, one log for validation. The reason for this was to avoid learning specific features of knots which might have affected the subsequent knot segmentation training.
The test set consisted in this case of all logs used for the subsequent knot segmentation training (set B), i.e. the various reconstructions were inferred from the models after finalised training.

The segmentation performance on the test sets from each trained U-Net were compared by calculating the Dice score of the knot labels, both in accumulated fashion for each log and in slice-wise fashion to study the development of the Dice score when progressing through knot groups. In addition, the contours of the inferred knot labels were compared.

\section{Results and Discussion}\label{sec:results}

\subsection{Validity of \acs{LPD} reconstructions}\label{subsec:PSRN_res}

We first evaluated the performance of the 2.5D~\ac{LPD} reconstructions with a varying number of source positions and varying number of consecutive slices.
The average \ac{PSNR} (of all slices from 3 entire test logs) for 2.5D~\ac{LPD} for 5, 7, 9, 11, 13 and 15 source positions and 2, 3, 4, 5, 7, and 9 consecutive slices are presented in Table \ref{tab:results1}. As expected, the greater the number of source positions, the greater the average \ac{PSNR}. However, adding more consecutive slices does not necessarily provide a better quality of the reconstructed images. As can be observed from Table \ref{tab:results1}, 3 and 2 consecutive slices for most of the considered source positions are providing the best quality of reconstructions measured by \ac{PSNR}.
This result can be explained by the fact that in cases where the number of consecutive slices are greater than 3, the initial and last slices in the sub-volume often do not anymore contain similar information to the neighboring slices. This is due to the relatively large distance between slices, in relation to the size of typical biological features, like knots. We expect a larger number of consecutive slices to be more beneficial for data with more slices per feature, i.e. a higher resolution in the z-direction.

\begin{table}[ht!]
    \centering
    \caption{Average \acs{PSNR} (of all slices from 3 entire test logs) for 2.5D~\ac{LPD} for varying number source positions and number of consecutive slices.}
    \label{tab:results1}
    \par\medskip
    \begin{tabular}{c *{6}E}
    & \multicolumn{6}{c}{\# consecutive slices} \\
    \cline{2-7}
    \multicolumn{1}{c}{\# src.\@ pos.} & \multicolumn{1}{c}{2} & \multicolumn{1}{c}{3} & \multicolumn{1}{c}{4} & \multicolumn{1}{c}{5} & \multicolumn{1}{c}{7} & \multicolumn{1}{c}{9} \\
    \hline
    \multicolumn{1}{c}{5} & 30.87 & 31.26 & 31.01 & 30.75 & 30.45 & 30.82 \\
    \multicolumn{1}{c}{7} & 32.41 & 32.46 & 31.97 & 32.08 & 32.23 & 32.07 \\
    \multicolumn{1}{c}{9} & 32.93 & 33.34 & 33.16 & 32.76 & 32.99 & 32.98 \\
    \multicolumn{1}{c}{11} & 33.48 & 33.85 & 33.87 & 33.26 & 33.35 & 33.50 \\
    \multicolumn{1}{c}{13} & 34.47 & 34.12 & 34.18 & 33.54 & 33.84 & 33.78 \\
    \multicolumn{1}{c}{15} & 34.96 & 34.84 & 34.80 & 34.47 & 34.60 & 34.50 \\
    \end{tabular}
\end{table}

Subsequently, we compared 2.5D~\ac{LPD} with plain \ac{2D} \ac{LPD}, which can be regarded as a state-of-the art \ac{2D} learned reconstruction method.
The comparison of 2.5D~\ac{LPD} with plain \ac{2D} \ac{LPD} is done for a varying number of source positions while fixing the number of consecutive slices to $3$, which in most cases performed best in the previous evaluation (Table \ref{tab:results1}).
A quantitative comparison of both methods is presented in Table \ref{tab:results2} and a qualitative comparison is given by Figure~\ref{fig:recs_with_knots} and Figure~\ref{fig:recs_without_knots}. For all considered numbers of source positions, 2.5D~\ac{LPD} outperformed 2D \ac{LPD} in terms of \ac{PSNR}.
However, if instead the the corresponding \ac{SSIM} is compared, then the trend is inverted. The \ac{SSIM} was computed using a Gaussian window of size $11 \times 11$ and similar results were obtained with different kernel sizes. Visual inspection of the reconstructed log slices with and without knots (Figure~\ref{fig:recs_with_knots} and \ref{fig:recs_without_knots}) clearly shows that 2.5D~\ac{LPD} is superior regarding reconstructing knots and the border between sapwood and heartwood. Since \ac{PSNR} is more suitable than \ac{SSIM} to assess pixel value accuracy, it seems to align better with the task at hand, which is the ability to reconstruct small details from sparse input data.

\begin{table}[ht!]
\setlength\tabcolsep{4.3pt}
    \centering
    \caption{Performance metrics (average for all slices from 3 entire test logs) for different source positions for various reconstruction methods. 
    The number of consecutive slices is $3$ for 2.5D~\ac{LPD} in all scenarios presented here.}
    \label{tab:results2}
    \par\medskip
    \begin{tabular}{l r r  r r r r r r r r r r }
    \# src.\@ pos. & \multicolumn{2}{c}{5} & \multicolumn{2}{c}{7}  & \multicolumn{2}{c}{9} & \multicolumn{2}{c}{11} & \multicolumn{2}{c}{13} & \multicolumn{2}{c}{15} \\
    \toprule
    \ac{LPD} & 2D  & 2.5D & 2D  & 2.5D & 2D & 2.5D & 2D & 2.5D & 2D & 2.5D & 2D & 2.5D  \\
    \hline
    \acs{PSNR} & 30.21 & 31.26 & 31.65 & 32.46 & 32.50 & 33.34 & 33.37 & 33.85 & 33.72 & 34.12 & 33.43 & 34.84 \\
    \ac{SSIM} & 0.99 & 0.90 & 0.99 & 0.91 & 0.99 & 0.91 & 0.99 & 0.91 & 0.99 & 0.92 & 0.99 & 0.92 \\
    \bottomrule
    \end{tabular}
\end{table}

\begin{figure}[ht!]
\centering
\setlength\tabcolsep{3pt}
	\begin{tabular}{cccc}
	Ground truth & &  2D \acs{LPD} & 2.5D~\ac{LPD} \\
	\includegraphics[trim={1.1cm 1.2cm 0.8cm 0.5cm},clip, width=3cm]{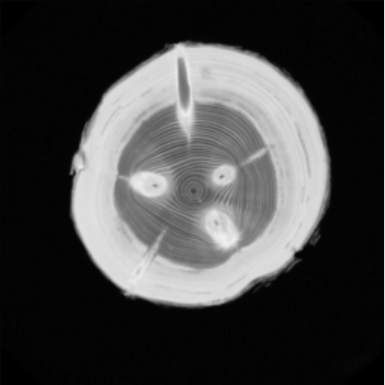} &
    $\quad$
    \put(-3,25){\rotatebox{90}{5 src.\@ pos.}} &
	\includegraphics[trim={1.1cm 1.2cm 0.8cm 0.5cm},clip, width=3cm]{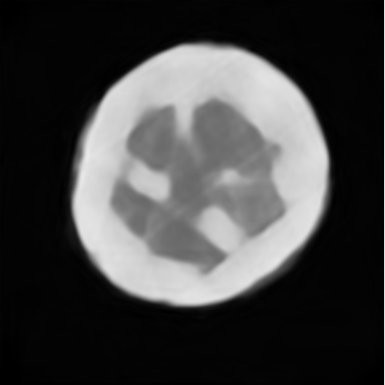} &
	\includegraphics[trim={1.1cm 1.2cm 0.8cm 0.5cm},clip, width=3cm]{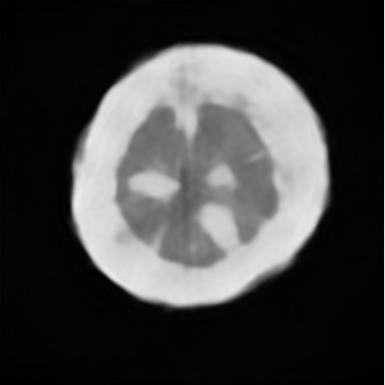} \\
    & & \acs{PSNR} 25.96 dB & \acs{PSNR} 27.34 dB \\
    \includegraphics[trim={1.1cm 1.2cm 0.8cm 0.5cm},clip, width=3cm]{ground_truth_log46_slice173.pdf} &
    $\quad$
    \put(-3,25){\rotatebox{90}{7 src.\@ pos.}} &	
    \includegraphics[trim={1.1cm 1.2cm 0.8cm 0.5cm},clip, width=3cm]{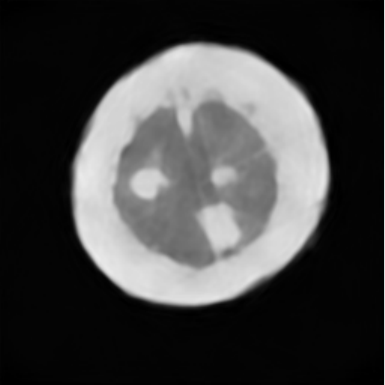} &
	\includegraphics[trim={1.1cm 1.2cm 0.8cm 0.5cm},clip, width=3cm]{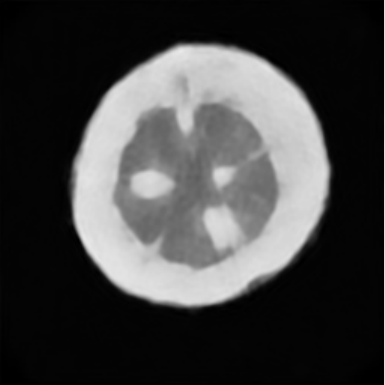} \\
    & & \acs{PSNR} 27.88 dB & \acs{PSNR} 29.03 dB \\
    \includegraphics[trim={1.1cm 1.2cm 0.8cm 0.5cm},clip, width=3cm]{ground_truth_log46_slice173.pdf} &
	$\quad$
    \put(-3,25){\rotatebox{90}{9 src.\@ pos.}} &
	\includegraphics[trim={1.1cm 1.2cm 0.8cm 0.5cm},clip, width=3cm]{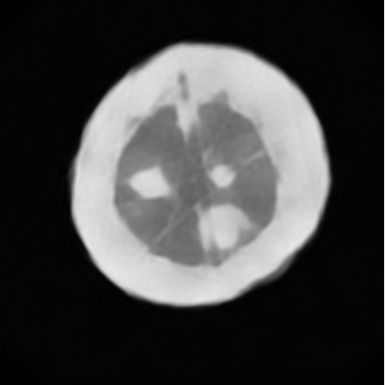} &
	\includegraphics[trim={1.1cm 1.2cm 0.8cm 0.5cm},clip, width=3cm]{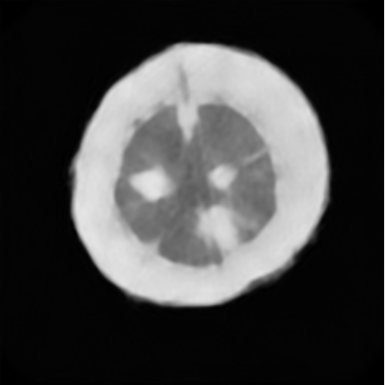} \\
    & & \acs{PSNR} 28.35 dB & \acs{PSNR} 29.52 dB \\
    \includegraphics[trim={1.1cm 1.2cm 0.8cm 0.5cm},clip, width=3cm]{ground_truth_log46_slice173.pdf} &
	$\quad$
    \put(-3,25){\rotatebox{90}{11 src.\@ pos.}} &
	\includegraphics[trim={1.1cm 1.2cm 0.8cm 0.5cm},clip, width=3cm]{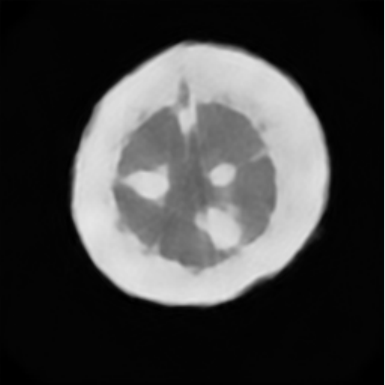} &
	\includegraphics[trim={1.1cm 1.2cm 0.8cm 0.5cm},clip, width=3cm]{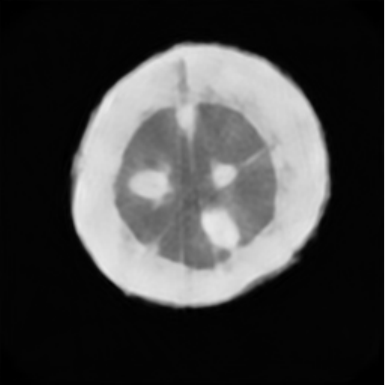} \\
    & & \acs{PSNR} 29.65 dB & \acs{PSNR} 29.80 dB \\
    \includegraphics[trim={1.1cm 1.2cm 0.8cm 0.5cm},clip, width=3cm]{ground_truth_log46_slice173.pdf} &
	$\quad$
    \put(-3,25){\rotatebox{90}{13 src.\@ pos.}} &
	\includegraphics[trim={1.1cm 1.2cm 0.8cm 0.5cm},clip, width=3cm]{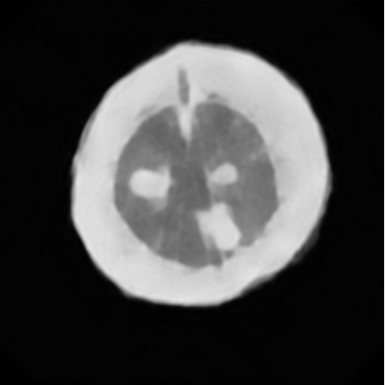} &
	\includegraphics[trim={1.1cm 1.2cm 0.8cm 0.5cm},clip, width=3cm]{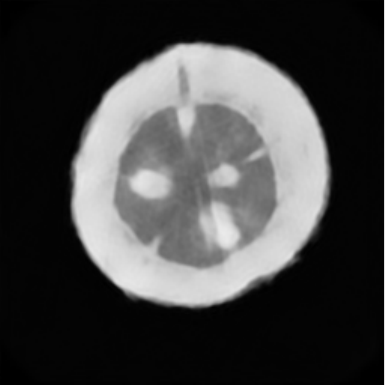} \\
    & & \acs{PSNR} 30.11 dB & \acs{PSNR} 30.56 dB \\
    \includegraphics[trim={1.1cm 1.2cm 0.8cm 0.5cm},clip, width=3cm]{ground_truth_log46_slice173.pdf} &
	$\quad$
    \put(-3,25){\rotatebox{90}{15 src.\@ pos.}} &
	\includegraphics[trim={1.1cm 1.2cm 0.8cm 0.5cm},clip, width=3cm]{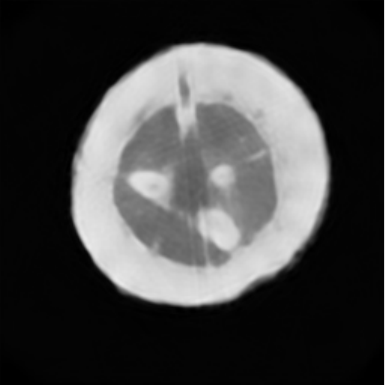} &
	\includegraphics[trim={1.1cm 1.2cm 0.8cm 0.5cm},clip, width=3cm]{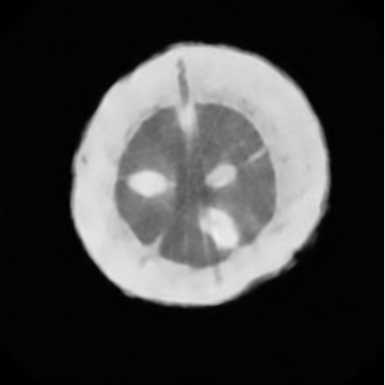} \\
    & & \acs{PSNR} 29.86 dB & \acs{PSNR} 31.41 dB \\
\end{tabular}
\caption{Reconstructions of a log slice with knots, using $3$ consecutive slices and varying number of source positions.}
\label{fig:recs_with_knots}
\end{figure}

\begin{figure}[ht!]
\centering
\setlength\tabcolsep{3pt}
	\begin{tabular}{cccc}
	Ground truth & & 2D \acs{LPD} & 2.5D~\ac{LPD} \\
	\includegraphics[trim={1.3cm 1.5cm 1.7cm 1.3cm},clip, width=3cm]{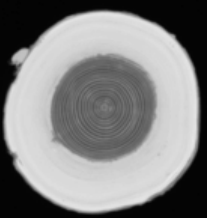} &
	$\quad$
    \put(-3,25){\rotatebox{90}{5 src.\@ pos.}} &
	\includegraphics[trim={1.3cm 1.5cm 1.7cm 1.3cm},clip, width=3cm]{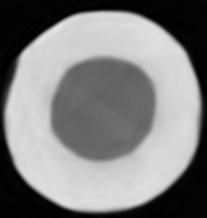} &
	\includegraphics[trim={1.3cm 1.5cm 1.7cm 1.3cm},clip, width=3cm]{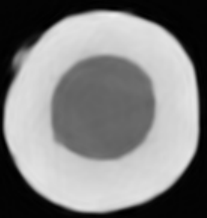} \\
    & & \acs{PSNR} 29.77 dB & \acs{PSNR} 34.10 dB \\
    \includegraphics[trim={1.3cm 1.5cm 1.7cm 1.3cm},clip, width=3cm]{ground_truth_log45_slice148.pdf} &
	$\quad$
    \put(-3,25){\rotatebox{90}{7 src.\@ pos.}} &
	\includegraphics[trim={1.3cm 1.5cm 1.7cm 1.3cm},clip, width=3cm]{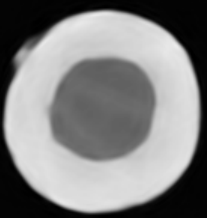} &
	\includegraphics[trim={1.3cm 1.5cm 1.7cm 1.3cm},clip, width=3cm]{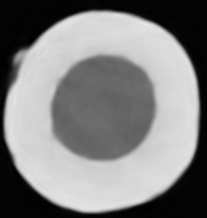} \\
    & & \acs{PSNR} 32.39 dB & \acs{PSNR} 34.80 dB \\
    \includegraphics[trim={1.3cm 1.5cm 1.7cm 1.3cm},clip, width=3cm]{ground_truth_log45_slice148.pdf} &
	$\quad$
    \put(-3,25){\rotatebox{90}{9 src.\@ pos.}} &
	\includegraphics[trim={1.3cm 1.5cm 1.7cm 1.3cm},clip, width=3cm]{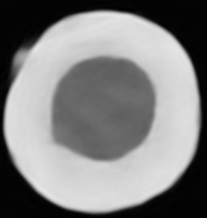} &
	\includegraphics[trim={1.3cm 1.5cm 1.7cm 1.3cm},clip, width=3cm]{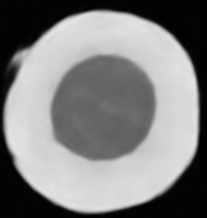} \\
    & & \acs{PSNR} 33.70 dB & \acs{PSNR} 35.79 dB \\
    \includegraphics[trim={1.3cm 1.5cm 1.7cm 1.3cm},clip, width=3cm]{ground_truth_log45_slice148.pdf} &
	$\quad$
    \put(-3,25){\rotatebox{90}{11 src.\@ pos.}} &
	\includegraphics[trim={1.3cm 1.5cm 1.7cm 1.3cm},clip, width=3cm]{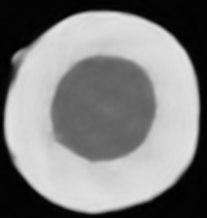} &
	\includegraphics[trim={1.3cm 1.5cm 1.7cm 1.3cm},clip, width=3cm]{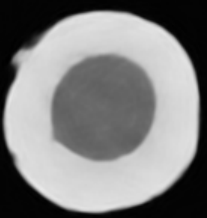} \\
    & & \acs{PSNR} 35.00 dB & \acs{PSNR} 36.12 dB \\
    \includegraphics[trim={1.3cm 1.5cm 1.7cm 1.3cm},clip, width=3cm]{ground_truth_log45_slice148.pdf} &
	$\quad$
    \put(-3,25){\rotatebox{90}{13 src.\@ pos.}} &
	\includegraphics[trim={1.3cm 1.5cm 1.7cm 1.3cm},clip, width=3cm]{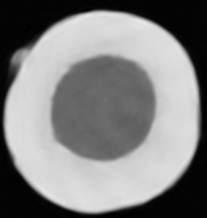} &
	\includegraphics[trim={1.3cm 1.5cm 1.7cm 1.3cm},clip, width=3cm]{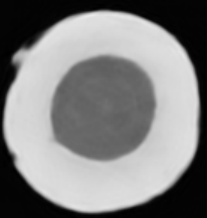} \\
    & & \acs{PSNR} 35.31 dB & \acs{PSNR} 36.73 dB \\
    \includegraphics[trim={1.3cm 1.5cm 1.7cm 1.3cm},clip, width=3cm]{ground_truth_log45_slice148.pdf} &
	$\quad$
    \put(-3,25){\rotatebox{90}{15 src.\@ pos.}} &
	\includegraphics[trim={1.3cm 1.5cm 1.7cm 1.3cm},clip, width=3cm]{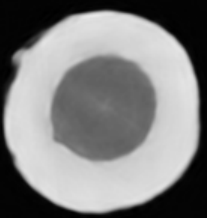} &
	\includegraphics[trim={1.3cm 1.5cm 1.7cm 1.3cm},clip, width=3cm]{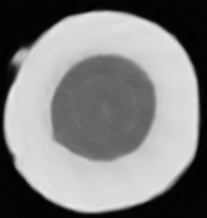} \\
    & & \acs{PSNR} 34.94 dB & \acs{PSNR} 37.45 dB \\
\end{tabular}
\caption{Reconstructions of a log slice without knots, using $3$ consecutive slices and varying number of source positions.}
\label{fig:recs_without_knots}
\end{figure}

\begin{figure}[ht!]
\centering
    \includegraphics[width=0.7\linewidth]{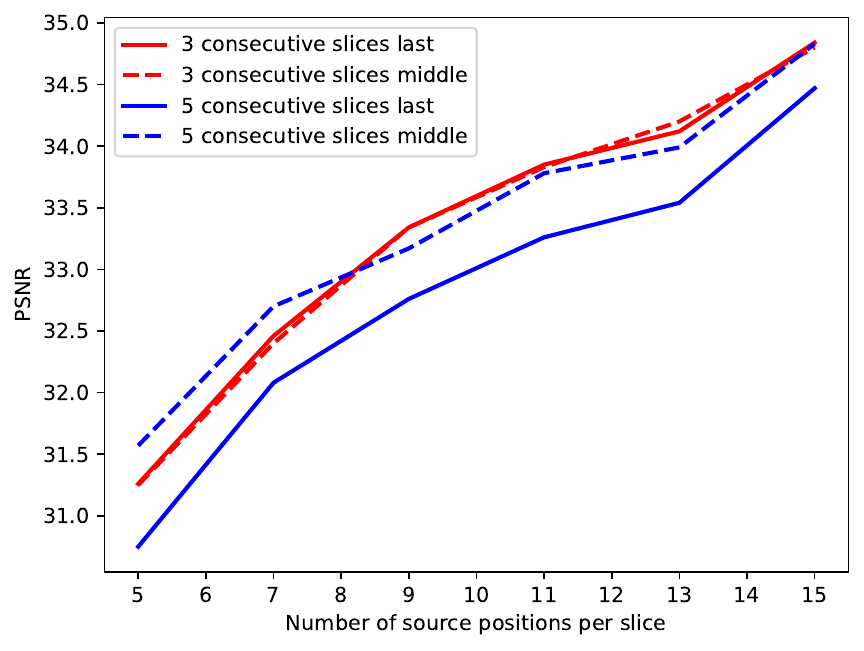}
\caption{Average \ac{PSNR} (of all slices from 3 entire test logs) for 2.5D~\ac{LPD} for different number of consecutive slices and different number of source positions per slice. ``Last'' and ``middle'' corresponds to two different strategies where last and middle slice is being reconstructed.}
\label{fig:PSNR_for_different_num_of_cons_slices_last_vs_middle}
\end{figure}

\begin{figure}[ht!]
\centering
\setlength\tabcolsep{3pt}
	\begin{tabular}{cccc}
	Ground truth & & Last & Middle \\
	\includegraphics[trim={0.9cm 0.9cm 0.7cm 0.4cm},clip, width=3cm]{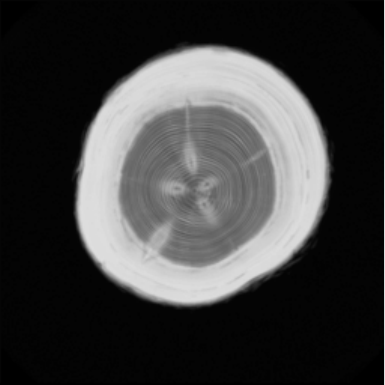} &
    $\quad$
    \put(-3,25){\rotatebox{90}{5 src.\@ pos.}} &
	\includegraphics[trim={0.9cm 0.9cm 0.7cm 0.4cm},clip, width=3cm]{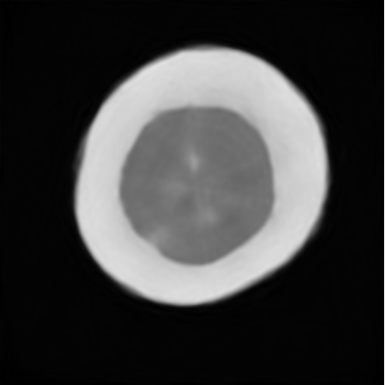} &
	\includegraphics[trim={0.9cm 0.9cm 0.7cm 0.4cm},clip, width=3cm]{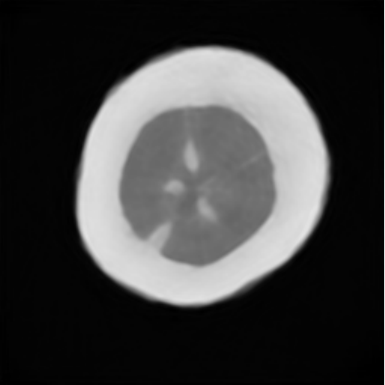} \\
    & & \ac{PSNR} 30.91 dB & \ac{PSNR} 33.00 dB \\
    \includegraphics[trim={0.9cm 0.9cm 0.7cm 0.4cm},clip, width=3cm]{ground_truth_log46_slice169.pdf} &
    $\quad$    
	\put(-3,25){\rotatebox{90}{7 src.\@ pos.}} &
	\includegraphics[trim={0.9cm 0.9cm 0.7cm 0.4cm},clip, width=3cm]{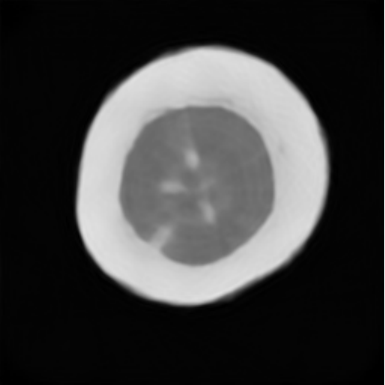} &
	\includegraphics[trim={0.9cm 0.9cm 0.7cm 0.4cm},clip, width=3cm]{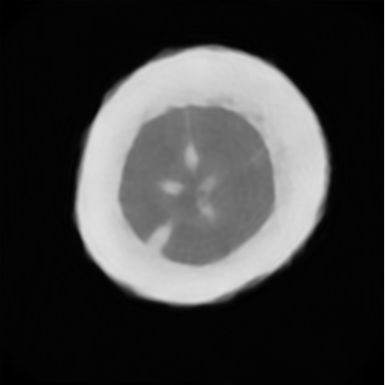} \\
    & & \ac{PSNR} 33.10 dB & \ac{PSNR} 33.40 dB \\
    \includegraphics[trim={0.9cm 0.9cm 0.7cm 0.4cm},clip, width=3cm]{ground_truth_log46_slice169.pdf} &
    $\quad$
	\put(-3,25){\rotatebox{90}{9 src.\@ pos.}} &
	\includegraphics[trim={0.9cm 0.9cm 0.7cm 0.4cm},clip, width=3cm]{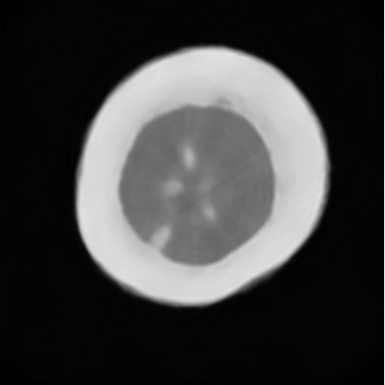} &
	\includegraphics[trim={0.9cm 0.9cm 0.7cm 0.4cm},clip, width=3cm]{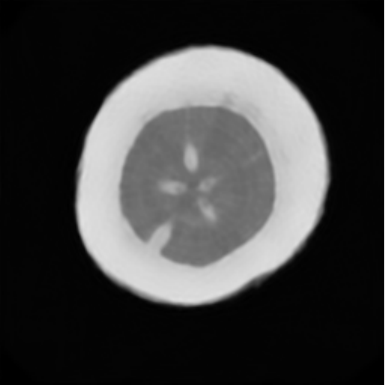} \\
    & & \ac{PSNR} 33.42 dB & \ac{PSNR} 34.56 dB \\
    \includegraphics[trim={0.9cm 0.9cm 0.7cm 0.4cm},clip, width=3cm]{ground_truth_log46_slice169.pdf} &
    $\quad$
	\put(-3,25){\rotatebox{90}{11 src.\@ pos.}} &
	\includegraphics[trim={0.9cm 0.9cm 0.7cm 0.4cm},clip, width=3cm]{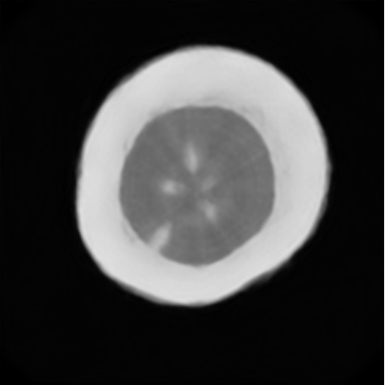} &
	\includegraphics[trim={0.9cm 0.9cm 0.7cm 0.4cm},clip, width=3cm]{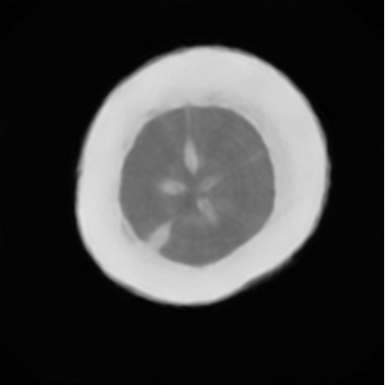} \\
    & & \ac{PSNR} 34.00 dB & \ac{PSNR} 35.08 dB \\
    \includegraphics[trim={0.9cm 0.9cm 0.7cm 0.4cm},clip, width=3cm]{ground_truth_log46_slice169.pdf} &
    $\quad$
	\put(-3,25){\rotatebox{90}{13 src.\@ pos.}} &
	\includegraphics[trim={0.9cm 0.9cm 0.7cm 0.4cm},clip, width=3cm]{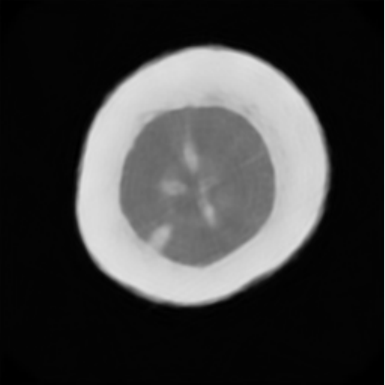} &
	\includegraphics[trim={0.9cm 0.9cm 0.7cm 0.4cm},clip, width=3cm]{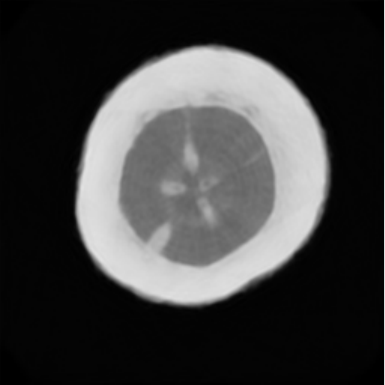} \\
    & & \ac{PSNR} 34.52 dB & \ac{PSNR} 35.11 dB \\
    \includegraphics[trim={0.9cm 0.9cm 0.7cm 0.4cm},clip, width=3cm]{ground_truth_log46_slice169.pdf} &
    $\quad$
	\put(-3,25){\rotatebox{90}{15 src.\@ pos.}} &
	\includegraphics[trim={0.9cm 0.9cm 0.7cm 0.4cm},clip, width=3cm]{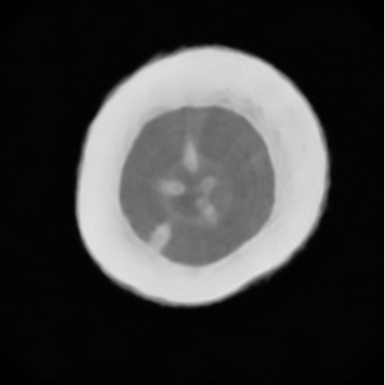} &
	\includegraphics[trim={0.9cm 0.9cm 0.7cm 0.4cm},clip, width=3cm]{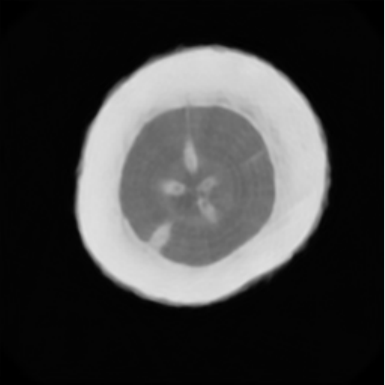} \\
    & & \ac{PSNR} 35.29 dB & \ac{PSNR} 36.06 dB \\
\end{tabular}
\caption{Reconstructions of log using $5$ consecutive slices and different number of source positions - sample where knots start appearing. ``Last'' and ``middle'' corresponds to two different strategies where last and middle slice is being reconstructed.}
\label{fig:last_vs_middle_knots_start_appearing}
\end{figure}

Finally, the two strategies (``last'' and ``middle'') for the 2.5D~\ac{LPD} reconstruction were compared. Both versions were evaluated for 3 and 5 consecutive slices, i.e. the best performing odd numbers from the previous evaluations (Table \ref{tab:results1}) to enable the ``middle'' strategy.
Figure~\ref{fig:PSNR_for_different_num_of_cons_slices_last_vs_middle} presents the average \ac{PSNR} of all slices from 3 entire test logs for these evaluations.
It can be seen that for 3 consecutive slices there is almost no difference between the ``last'' and ``middle'' strategy while for 5 consecutive slices the ``middle'' strategy outperforms the ``last'' strategy. Our interpretation is that in cases where a slice just contains the start of a knot, the ``middle'' strategy performs better in comparison to ``last''. This happens because the ``middle'' strategy also sees several of the subsequent slices which contain information on the same knot group, and in that way the model can anticipate the appearance of a knot.
In contrast, the ``last'' strategy works in retrospect, i.e. it sees some preceding slices which do not contain information on knots and it is therefore less likely to successfully reconstruct those parts of the images.
This is demonstrated in Figure~\ref{fig:last_vs_middle_knots_start_appearing} where one such slice is reconstructed by both strategies and its reconstructions from a different number of source positions are presented alongside with the ground truth slice.
The plotted curves for 5 to 7 source positions in Figure~\ref{fig:PSNR_for_different_num_of_cons_slices_last_vs_middle} support this finding: for those small numbers of source positions, which are the most interesting in terms of industrial applicability, the ``middle'' strategy with 5 consecutive slices outperforms the other strategies. 
Figure~\ref{fig:last_vs_middle_knots_start_appearing} shows that the main benefit of the ``middle'' strategy lies in the reconstruction of the beginning of a knot group, whereas the ``last'' strategy delivers clearly worse results. Hence, for 5 and 7 source positions the ``middle'' instead of the ``last'' strategy should be preferred, while for more source positions the richer data outweighs any benefits of the ``middle'' strategy over the ``last'' strategy.
These results motivated our choice of \ac{LPD} candidates for the evaluation of the suitability for knot segmentation in the next section.

\subsection{Knot segmentation performance}
Table \ref{tab:unet-regions-dices} shows the Dice scores for all segmentation scenarios in this study. It shows a slice-wise evaluation for the first 20\% (Start), the following 60\% (Mid) and the last 20\% (End) along the lengthwise distance of knot groups, the slice-wise mean for all knot groups, and the 3D total bulk Dice score.
For the slice-wise and 3D totals, the differences among scores for the learned reconstructions are small for the same number of source positions. The U-Nets trained on reconstructions from 9 source positions provide nearly the same segmentation performance as the U-Net trained on the full CT images, while the performance drops by approximately 15\% for U-Nets trained on 5 source positions. 

\begin{table}
\caption{Mean slice-wise Dice scores for knot segmentation evaluated at various positions within knot groups on test and validation datasets; the first 20\% (Start), the following 60\% (Mid) and the last 20\% (End) along the lengthwise distance of knot groups, and the slice-wise mean for all knot groups.
The lower part shows the full 3D bulk mean Dice scores for the knot segmentation. 
Results are compared across U-Net models trained on different reconstruction methods.
Note that in 2D slice-wise Dice scores, each slice is weighted equally, unlike full 3D Dice scores that emphasise slices with larger segmented areas, which results in different overall Dice scores due to the non-linear scaling behaviour of the Dice metric and the influence of empty or nearly empty slices.}
\label{tab:unet-regions-dices}
\par\medskip
\renewcommand{\arraystretch}{1} 
\begin{tabular}{llrrrrrrr}
& & & 
\multicolumn{3}{c}{\textbf{\small 9 src. pos.}} &
\multicolumn{3}{c}{\textbf{\small 5 src. pos.}} \\
\cmidrule(lr){4-6} \cmidrule(lr){7-9}
 & & 
\multicolumn{1}{c}{\textbf{\small CT}} & 
\makecell[r]{\textbf{\footnotesize 2.5D LPD,} \\[-7pt]
\textbf{\footnotesize 3 sl. mid}} & 
\textbf{\footnotesize 2D LPD} & 
\makecell[r]{\textbf{\footnotesize FBP \&} \\[-7pt]
\textbf{\footnotesize 2D UNet}} &
\makecell[r]{\textbf{\footnotesize 2.5D LPD,} \\[-7pt]
\textbf{\footnotesize 5 sl. last}} & 
\textbf{\footnotesize 2D LPD} &
\makecell[r]{\textbf{\footnotesize FBP \&} \\[-7pt]
\textbf{\footnotesize 2D UNet}} \\
\midrule
\multirow[t]{4}{*}{\textbf{Test}} 
& \textbf{Start} & 0.542 & 0.480 & 0.489 & 0.473 & 0.421 & 0.359 & 0.322 \\
& \textbf{Mid} & 0.785 & 0.736 & 0.760 & 0.742 & 0.695 & 0.687 & 0.677 \\
& \textbf{End} & 0.588 & 0.554 & 0.578 & 0.560 & 0.460 & 0.516 & 0.509 \\
& \textbf{Total} & 0.696 & 0.648 & 0.669 & 0.651 & 0.592 & 0.587 & 0.572 \\
\cline{1-9}
\multirow[t]{4}{*}{\textbf{Val}} 
& \textbf{Start} & 0.664 & 0.594 & 0.587 & 0.595 & 0.470 & 0.457 & 0.456 \\
& \textbf{Mid} & 0.806 & 0.765 & 0.767 & 0.760 & 0.712 & 0.680 & 0.699 \\
& \textbf{End} & 0.679 & 0.633 & 0.665 & 0.639 & 0.554 & 0.578 & 0.575 \\
& \textbf{Total} & 0.752 & 0.703 & 0.710 & 0.702 & 0.631 & 0.615 & 0.625 \\
\cline{1-9}
\textbf{Test} & \multirow[c]{2}{*}{\textbf{3D}} & 0.771 & 0.729 & 0.747 & 0.732 & 0.686 & 0.679 & 0.677 \\
\textbf{Val} & & 0.781 & 0.738 & 0.740 & 0.734 & 0.675 & 0.651 & 0.681 \\
\bottomrule
\end{tabular}
\end{table}

When evaluating the segmentation performance of the U-Nets on a per-slice basis for various regions in the knot groups, the results reveal a more nuanced picture.
The start of a knot group (when branches originate from the central pith in the tree located in the dry heartwood) is somewhat better detected in 2.5D~\ac{LPD} reconstructions than in the others for 5 source positions, potentially because of the chosen ``last'' strategy which can more efficiently anticipate incoming changes in the logs. 
For the Mid and End regions of knot groups, the results are similar, however, a slight advantage of \ac{LPD} based reconstructions compared to \ac{FBP} \& 2D U-Net based ones can be observed for fewer source positions, potentially due to the more effective utilisation of sparse information.
Note that the contours at the end the knot groups were challenging to detect for the human labellers, which makes interpretation of differences of scores in the End regions uncertain.

\begin{figure}[ht!]
\centering
\setlength\tabcolsep{0.5pt}
	\begin{tabular}{c}
	\includegraphics[width=9cm]{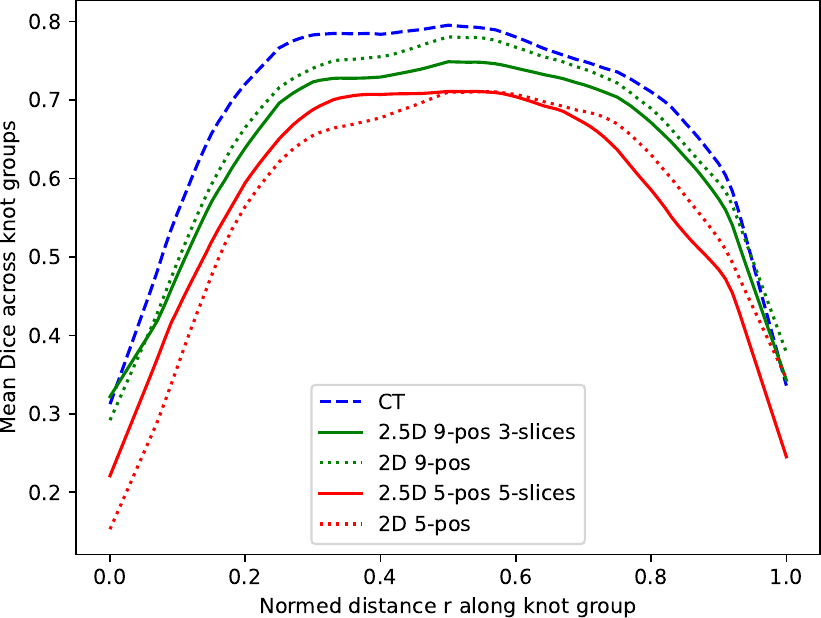} \\
        (a) \\
	\includegraphics[width=9cm]{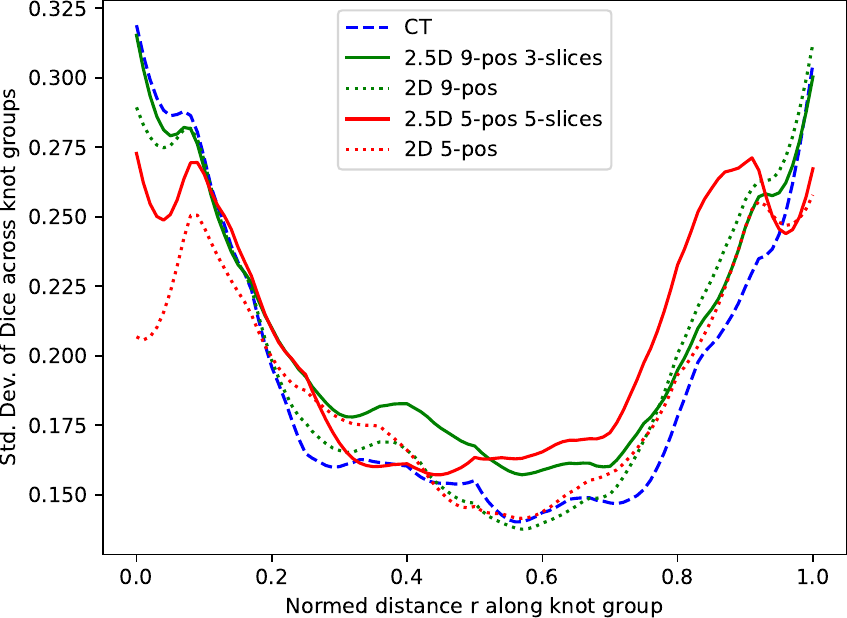} \\
        (b)
\end{tabular}
\caption{Test set mean slice-wise Dice scores (a) along a normalised distance $r$ from the start to the end of knot groups and the corresponding standard deviations of the scores (b), for the human labels (GT), the CT trained U-Net (CT) and the four \ac{LPD} trained U-Nets (2.5D 9-pos 3-slices, 2D 9-pos, 2.5D 5-pos 5-slices, and 2D 5-pos respectively).}
\label{fig:diceprofiles}
\end{figure}

\begin{figure}[ht!]
\centering
\setlength\tabcolsep{0.5pt}
	\begin{tabular}{c}
	\includegraphics[width=9cm]{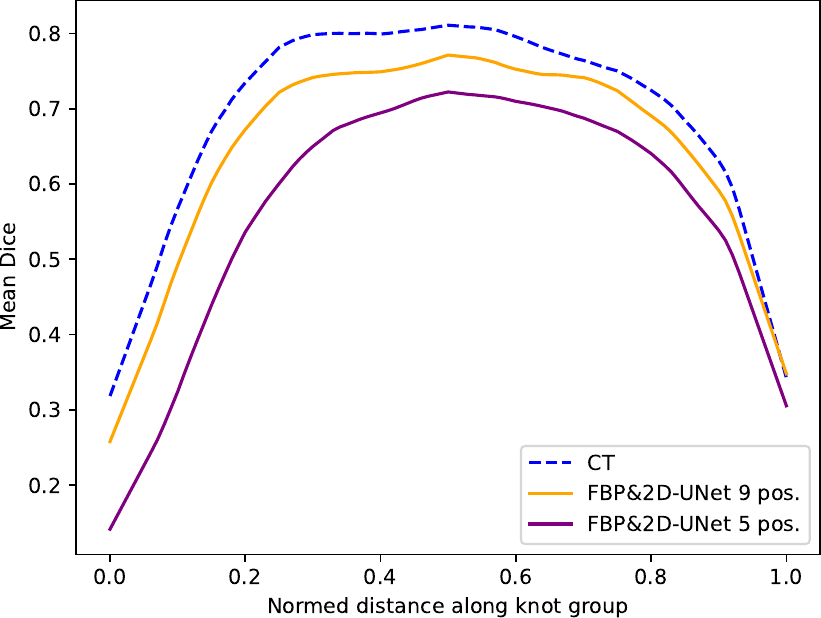} \\
        (a) \\
	\includegraphics[width=9cm]{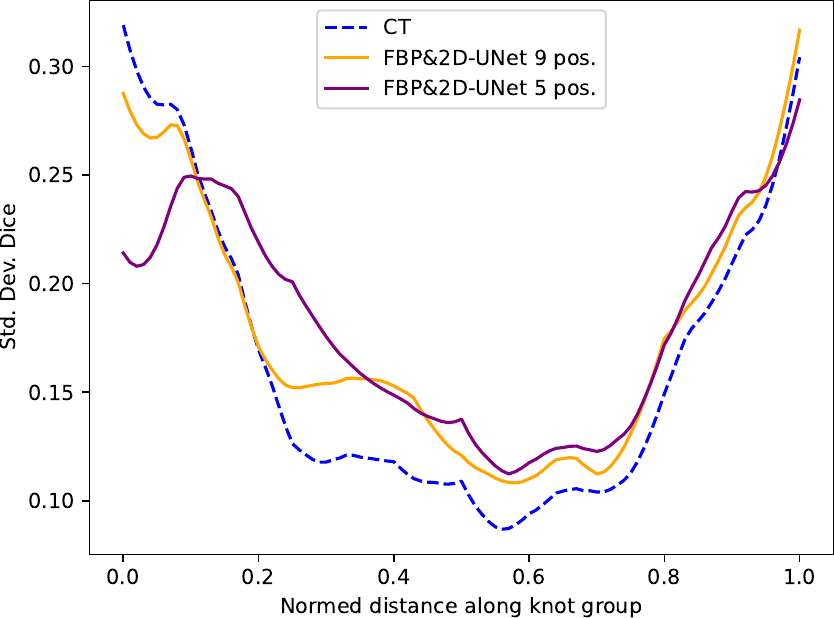} \\
        (b)
\end{tabular}
\caption{Test set mean slice-wise Dice scores (a) along a normalised distance $r$ from the start to the end of knot groups and the corresponding standard deviations of the scores (b), for the CT trained U-Net (CT) and the two FBP\&2D U-Nets (9 and 5 source positions).}
\label{fig:diceprofiles_FBP}
\end{figure}

\begin{figure}[ht!]
\centering
\setlength\tabcolsep{0.5pt}
	\begin{tabular}{ccc}
	CT and best 2D \acs{LPD} & 9 src.\@ pos. & 5 src.\@ pos. \\
	\includegraphics[width=4cm]{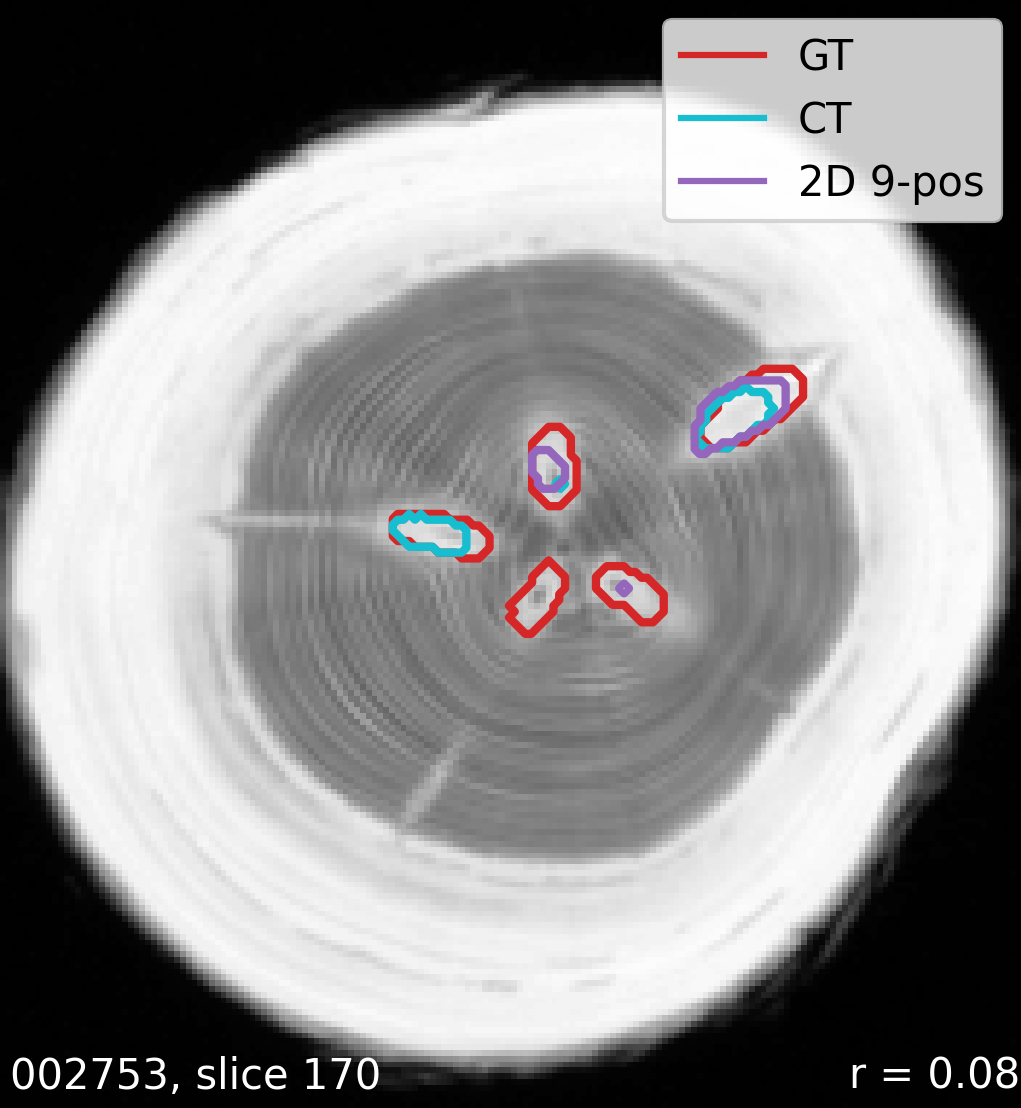} &
	\includegraphics[width=4cm]{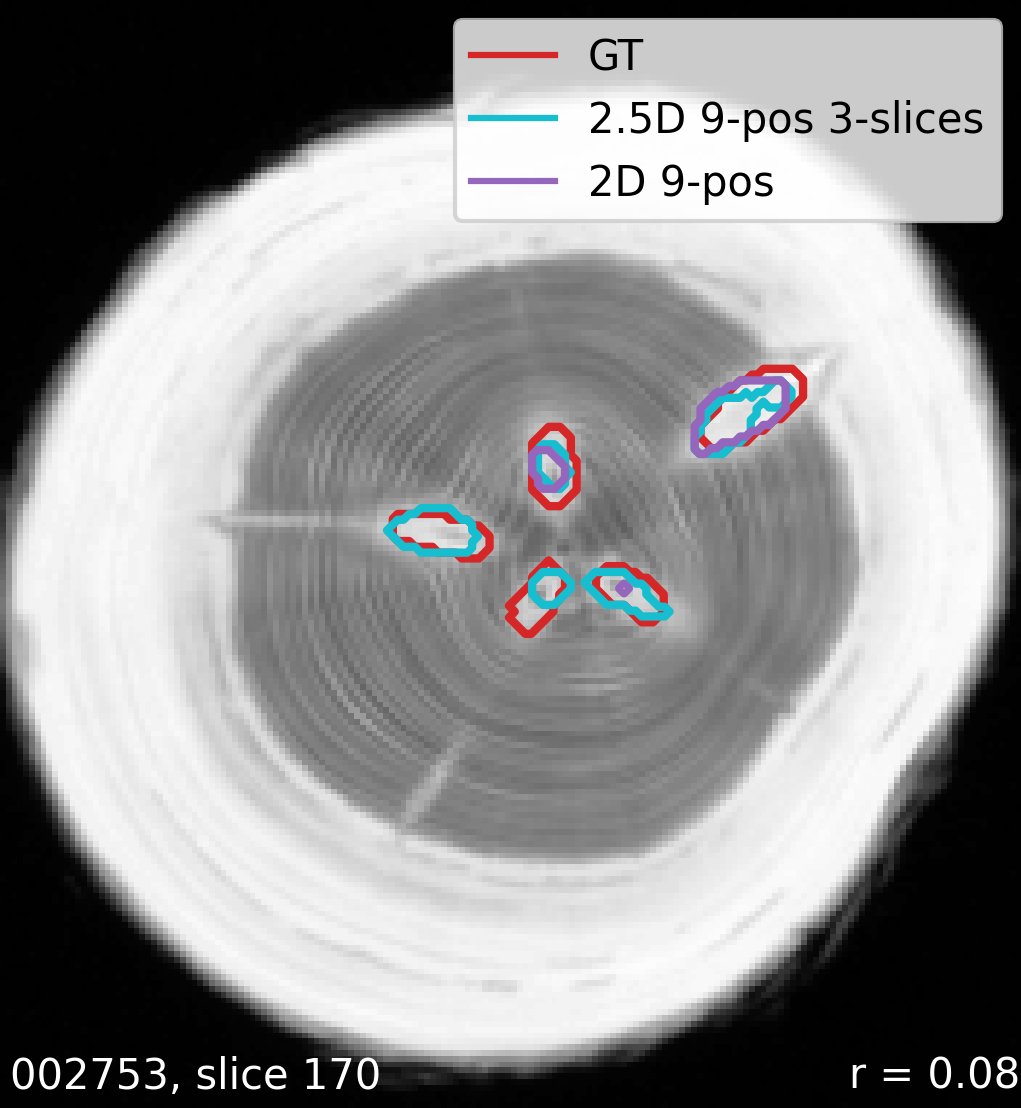} &
	\includegraphics[width=4cm]{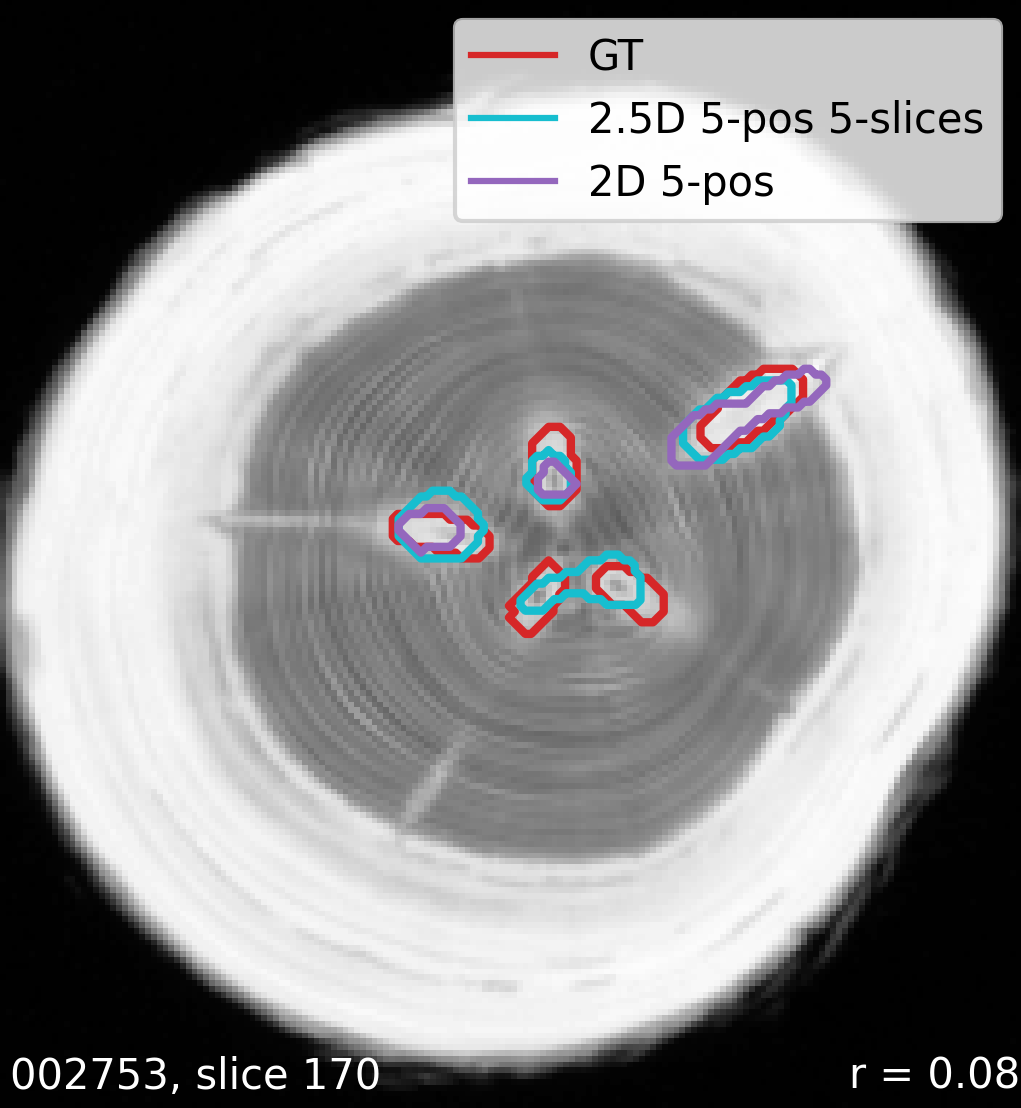} \\
        \includegraphics[width=4cm]{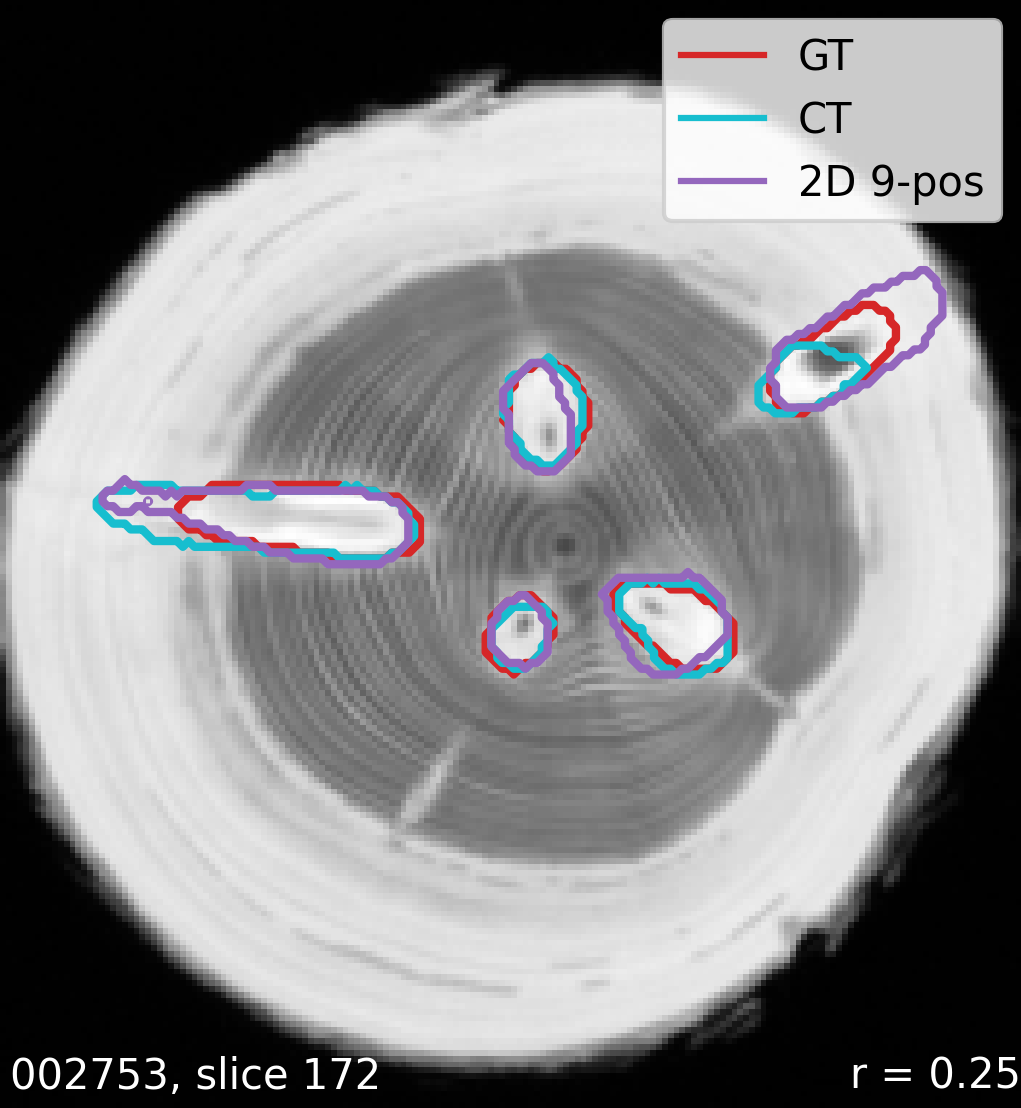} &
	\includegraphics[width=4cm]{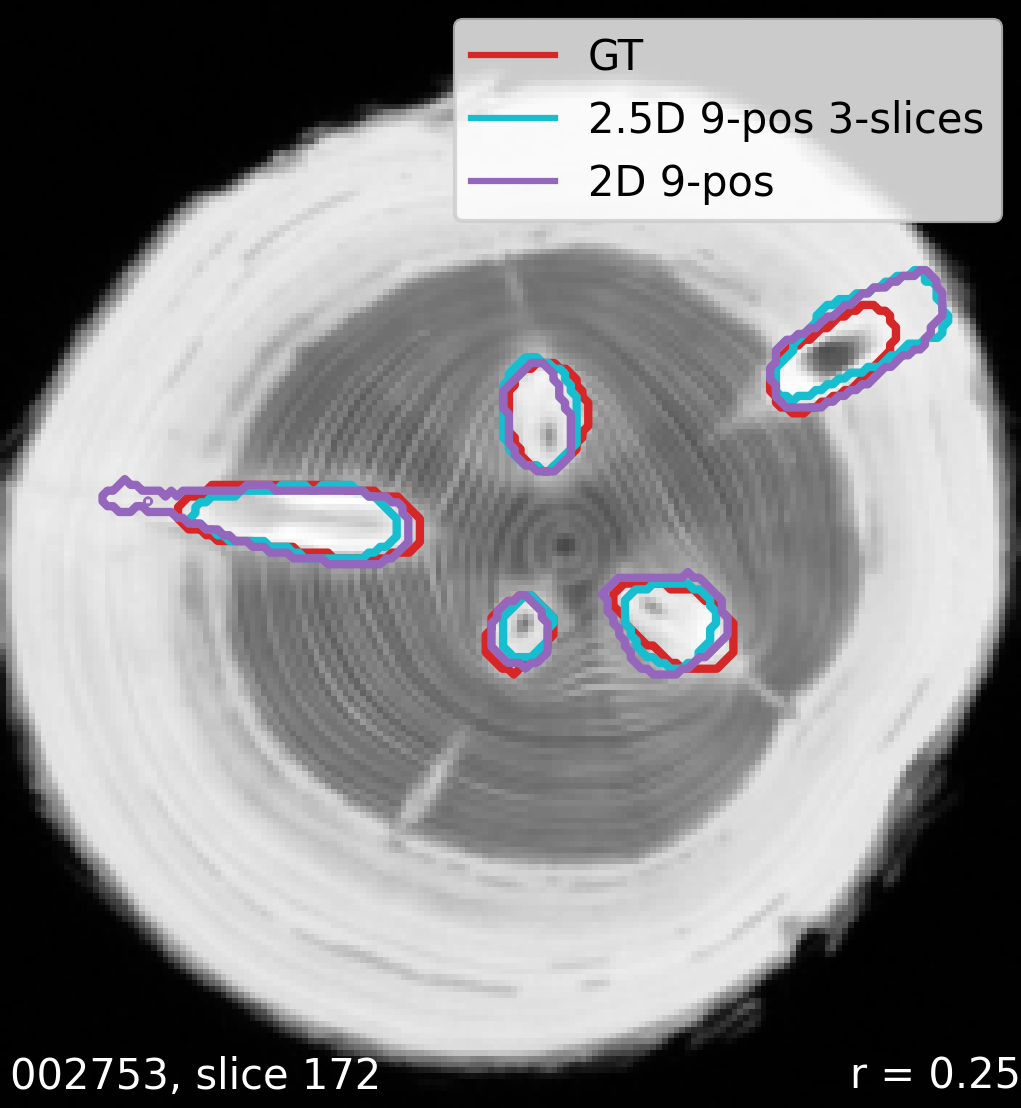} &
	\includegraphics[width=4cm]{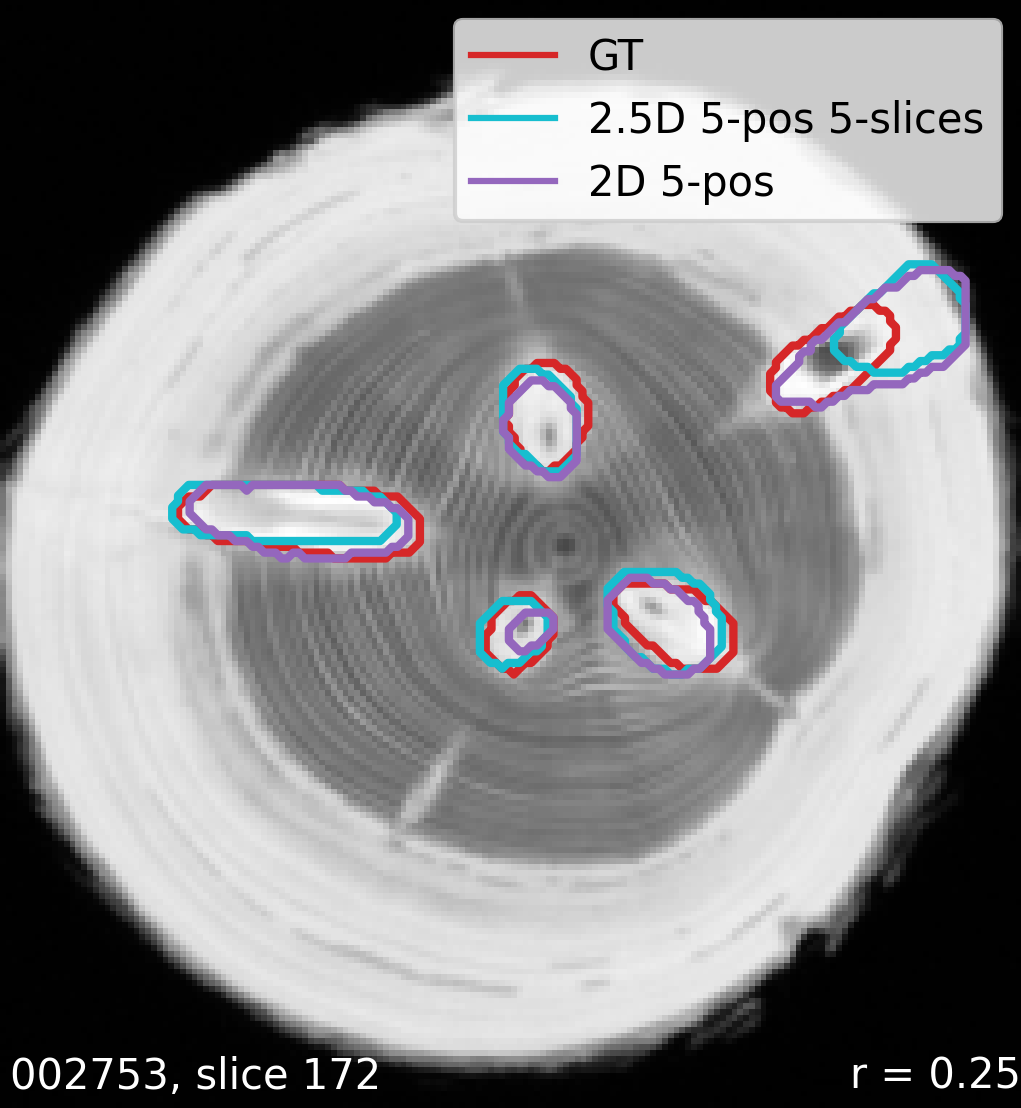} \\
        \includegraphics[width=4cm]{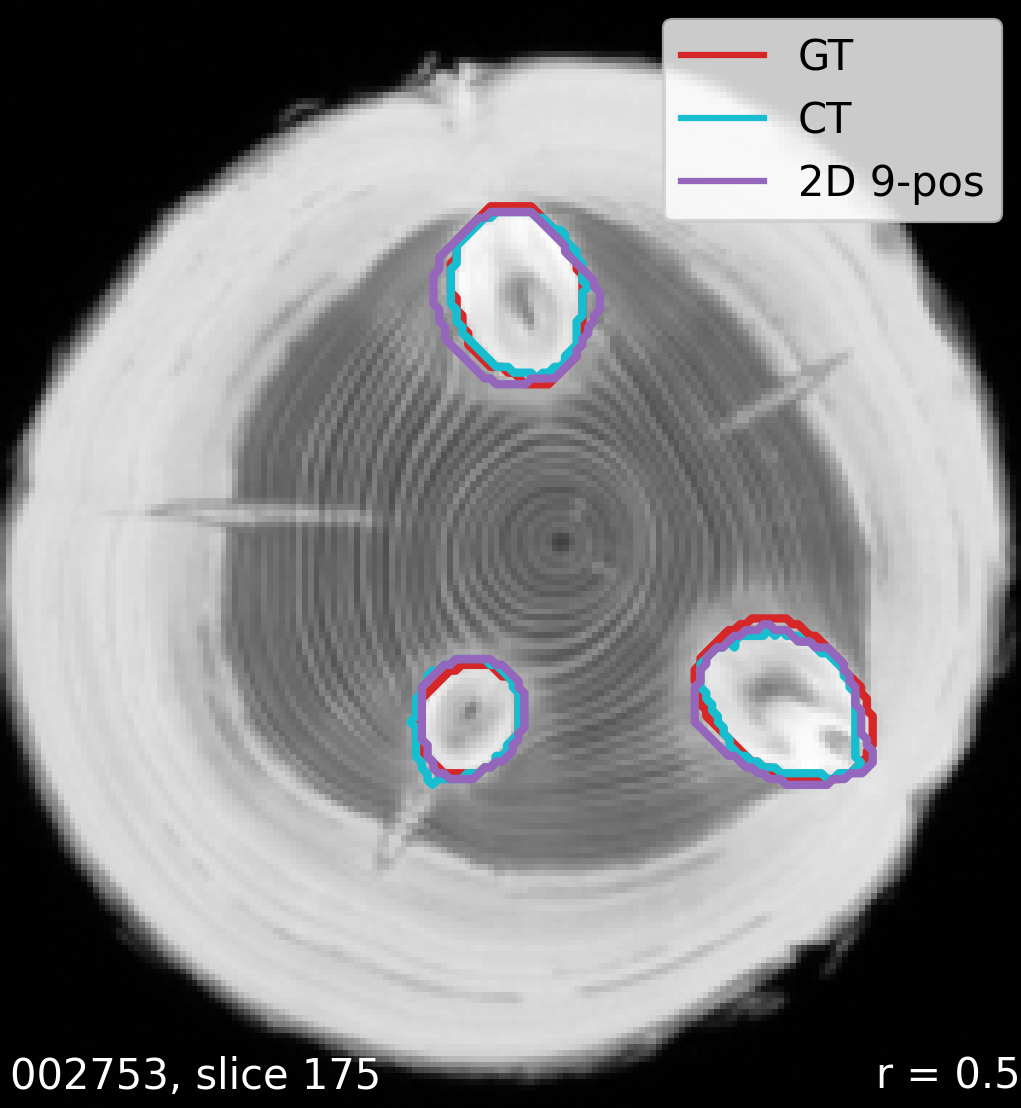} &
	\includegraphics[width=4cm]{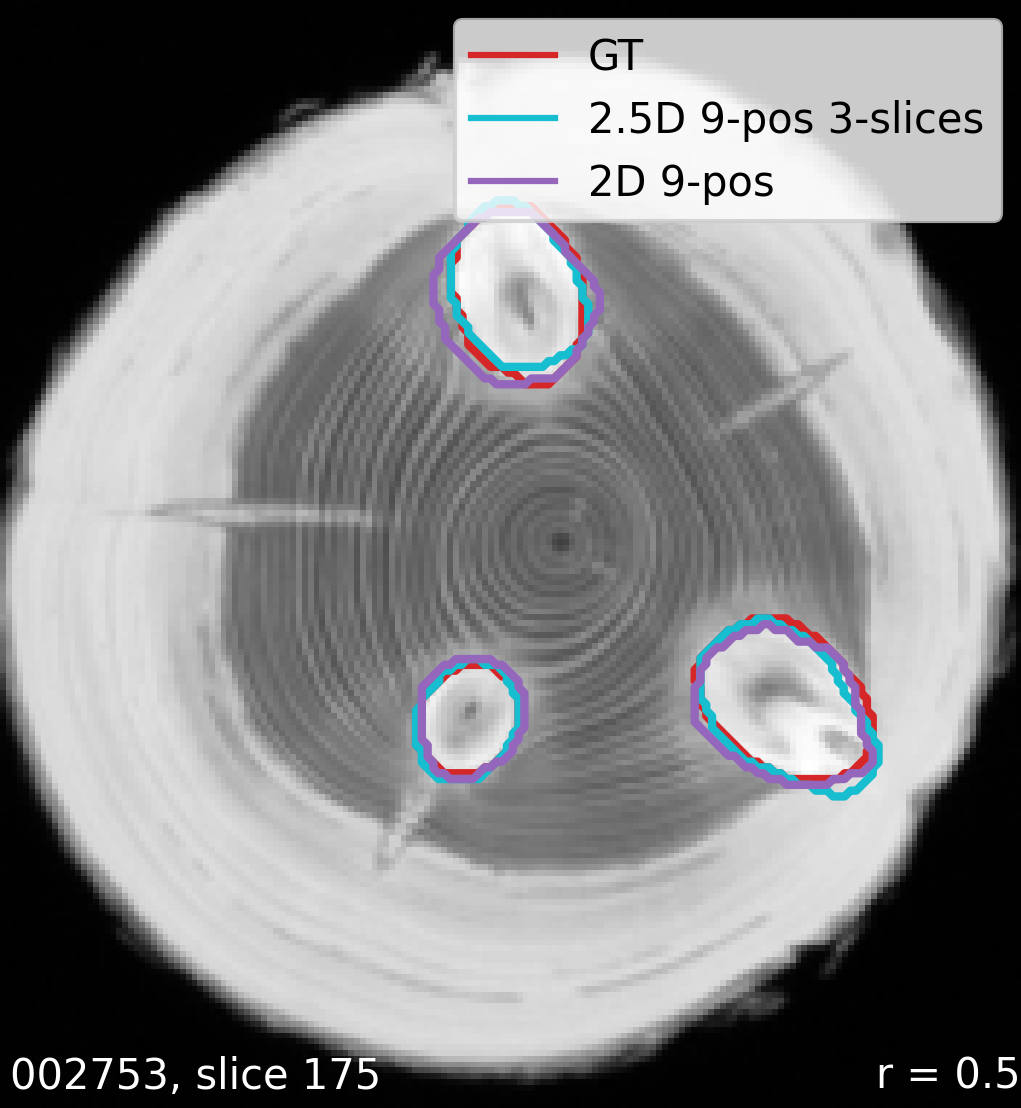} &
	\includegraphics[width=4cm]{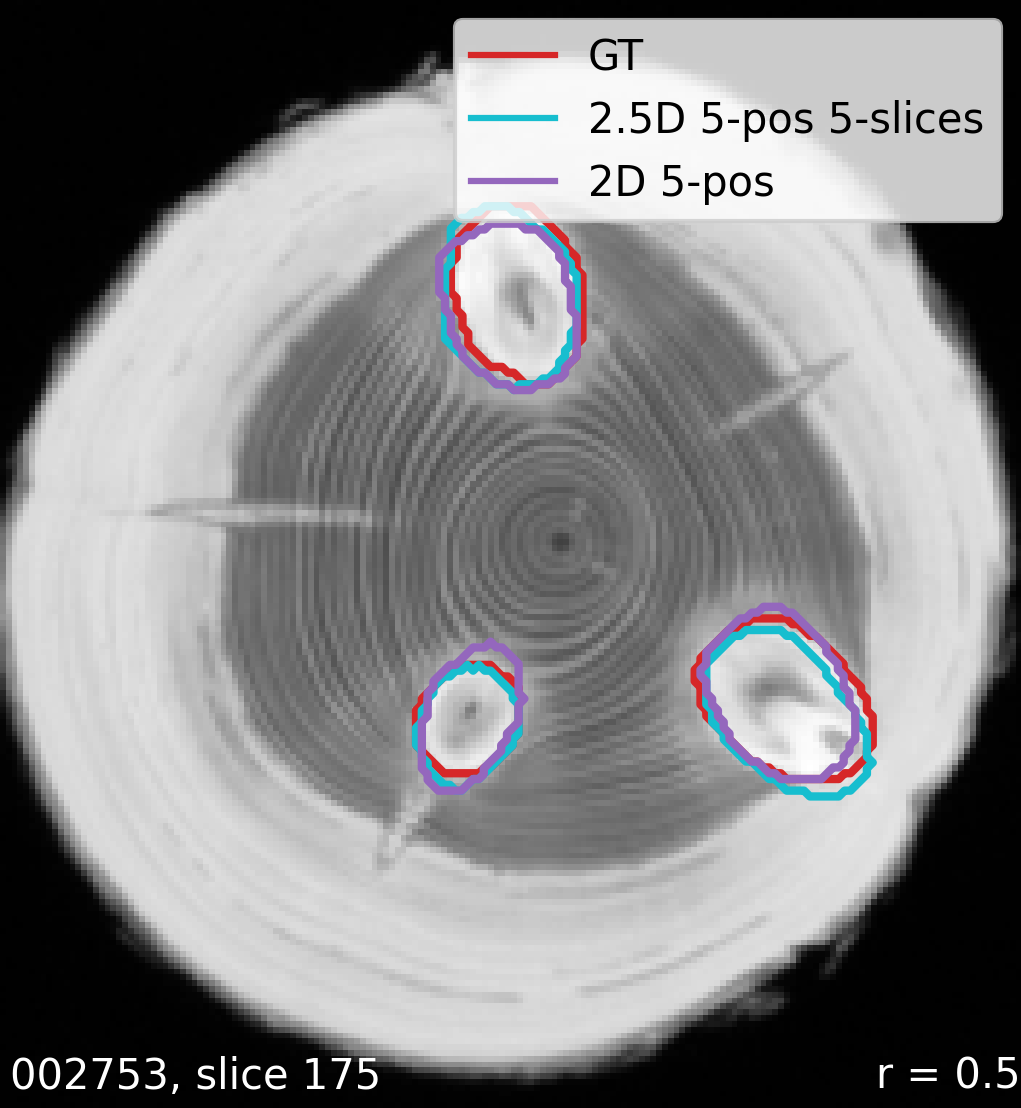} \\
        \includegraphics[width=4cm]{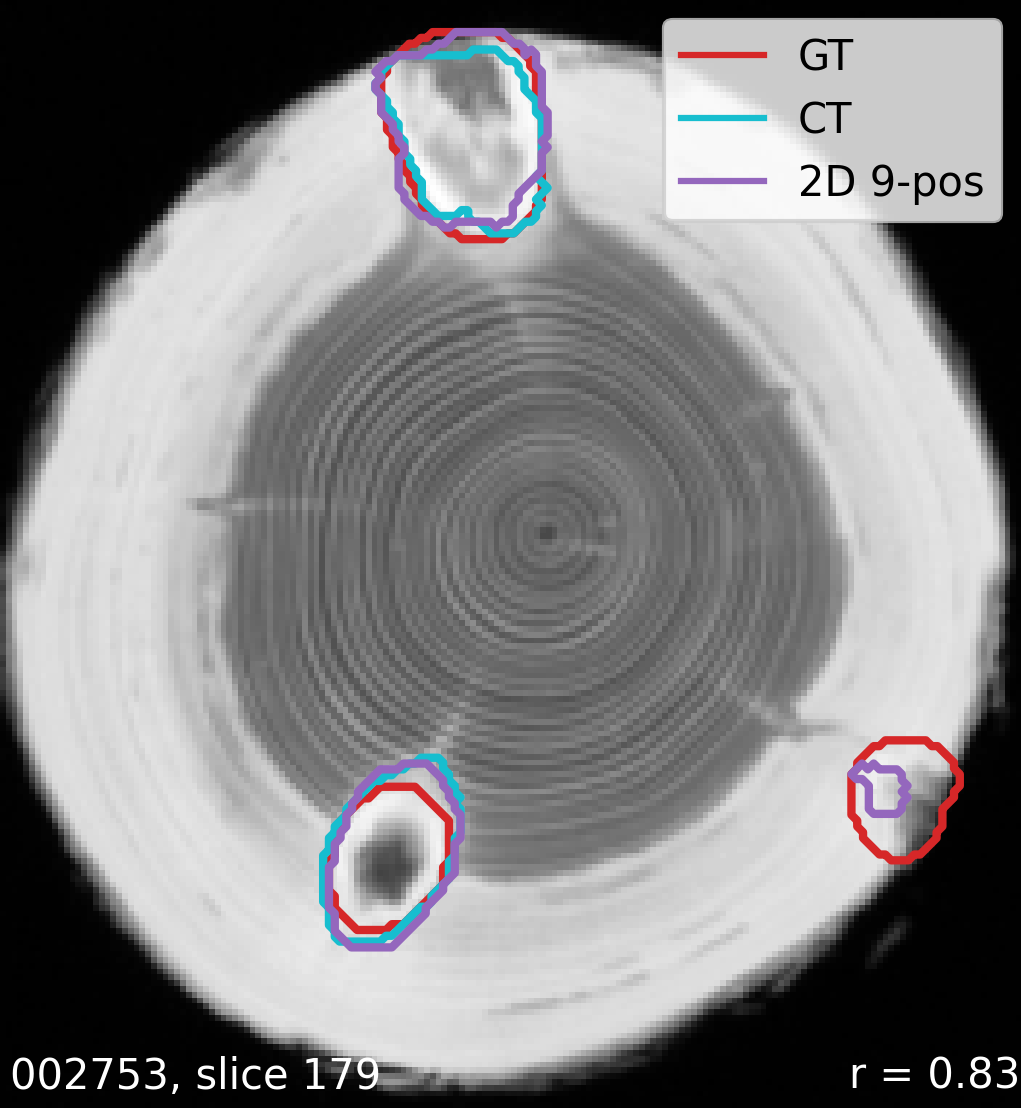} &
	\includegraphics[width=4cm]{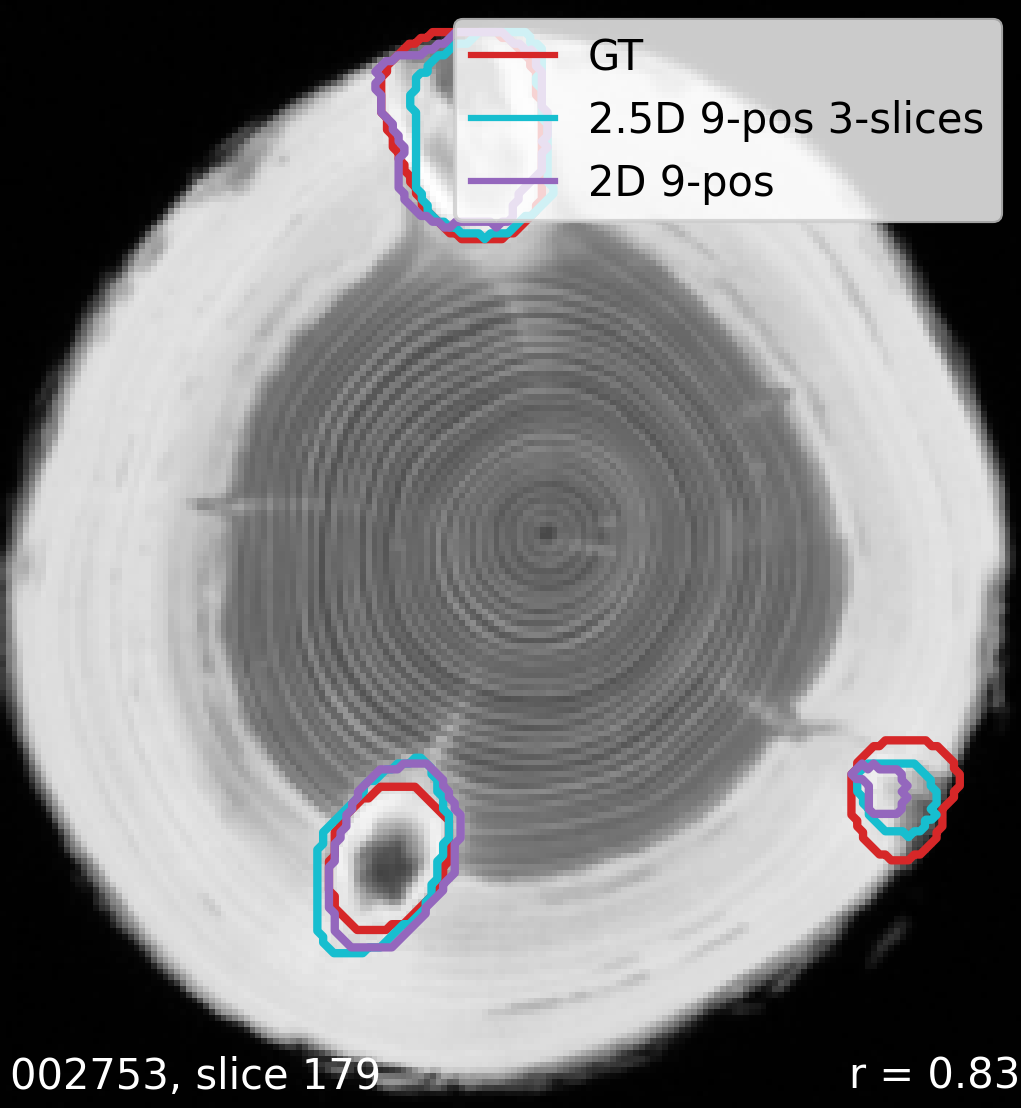} &
	\includegraphics[width=4cm]{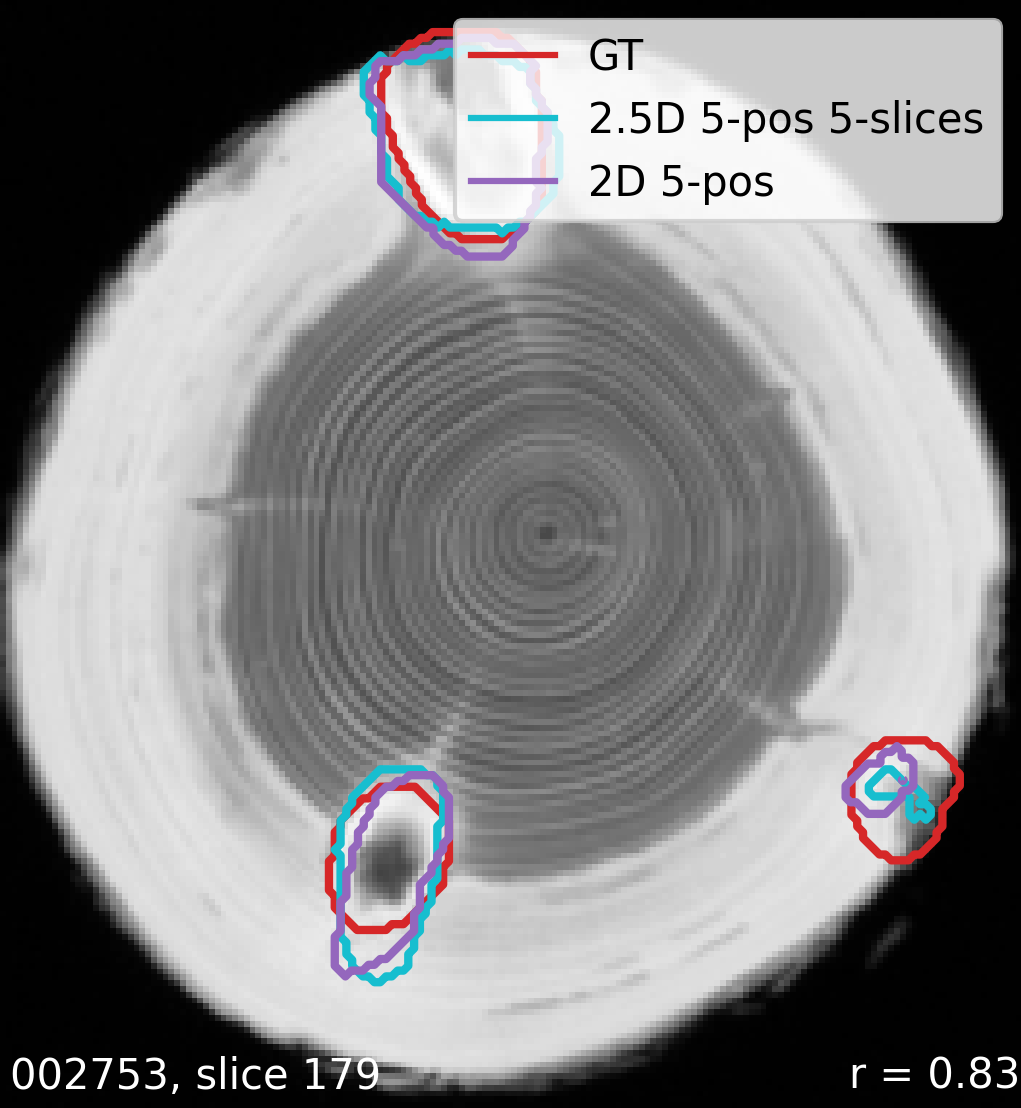} \\
        \includegraphics[width=4cm]{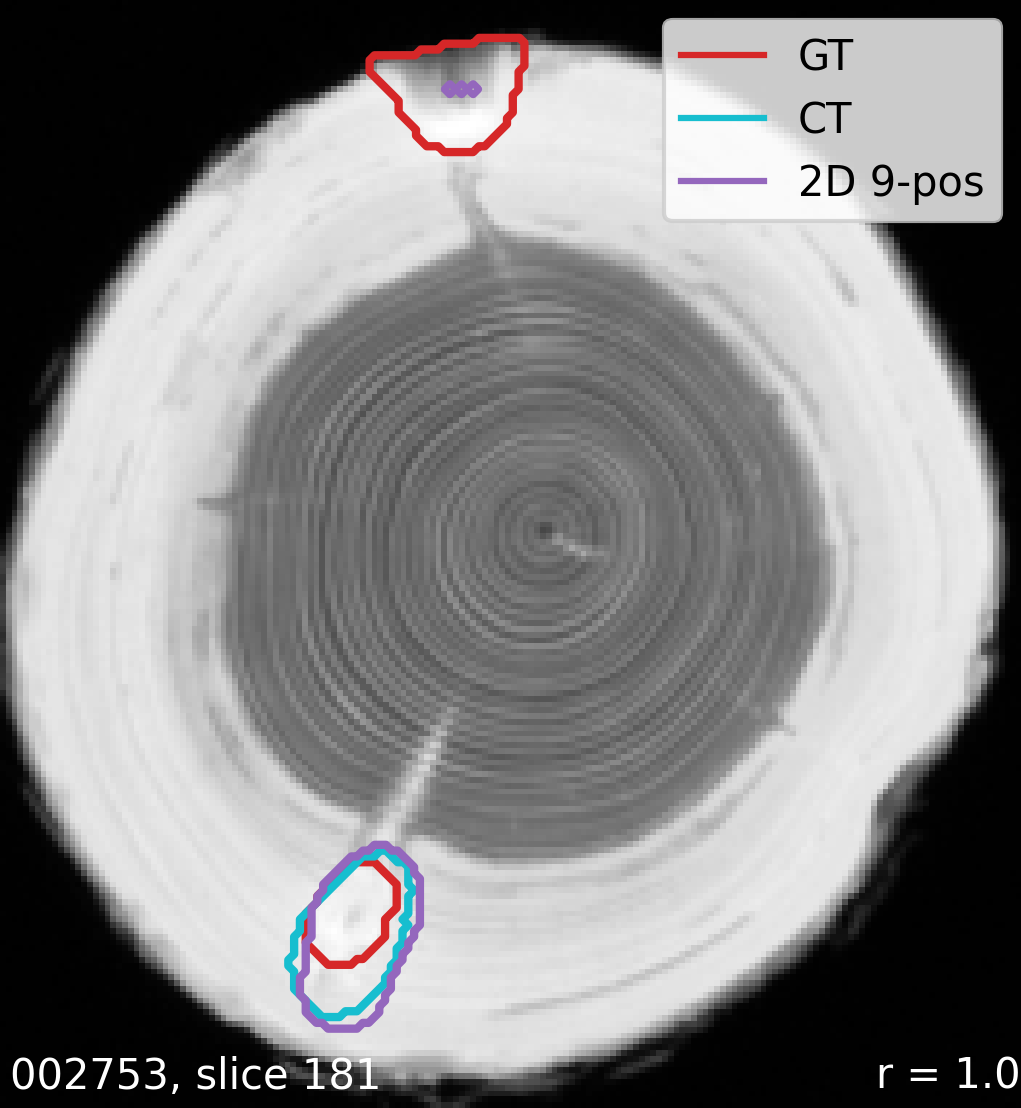} &
	\includegraphics[width=4cm]{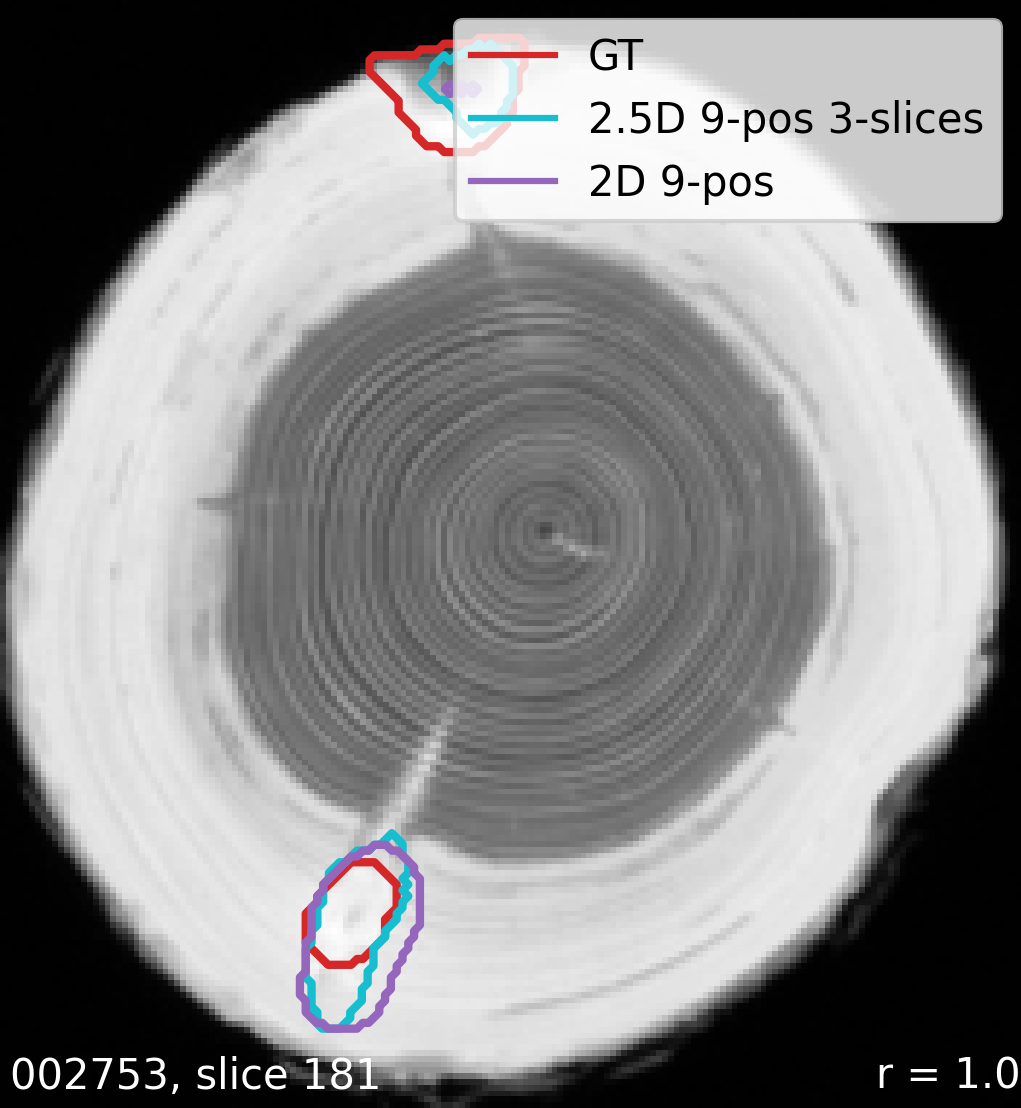} &
	\includegraphics[width=4cm]{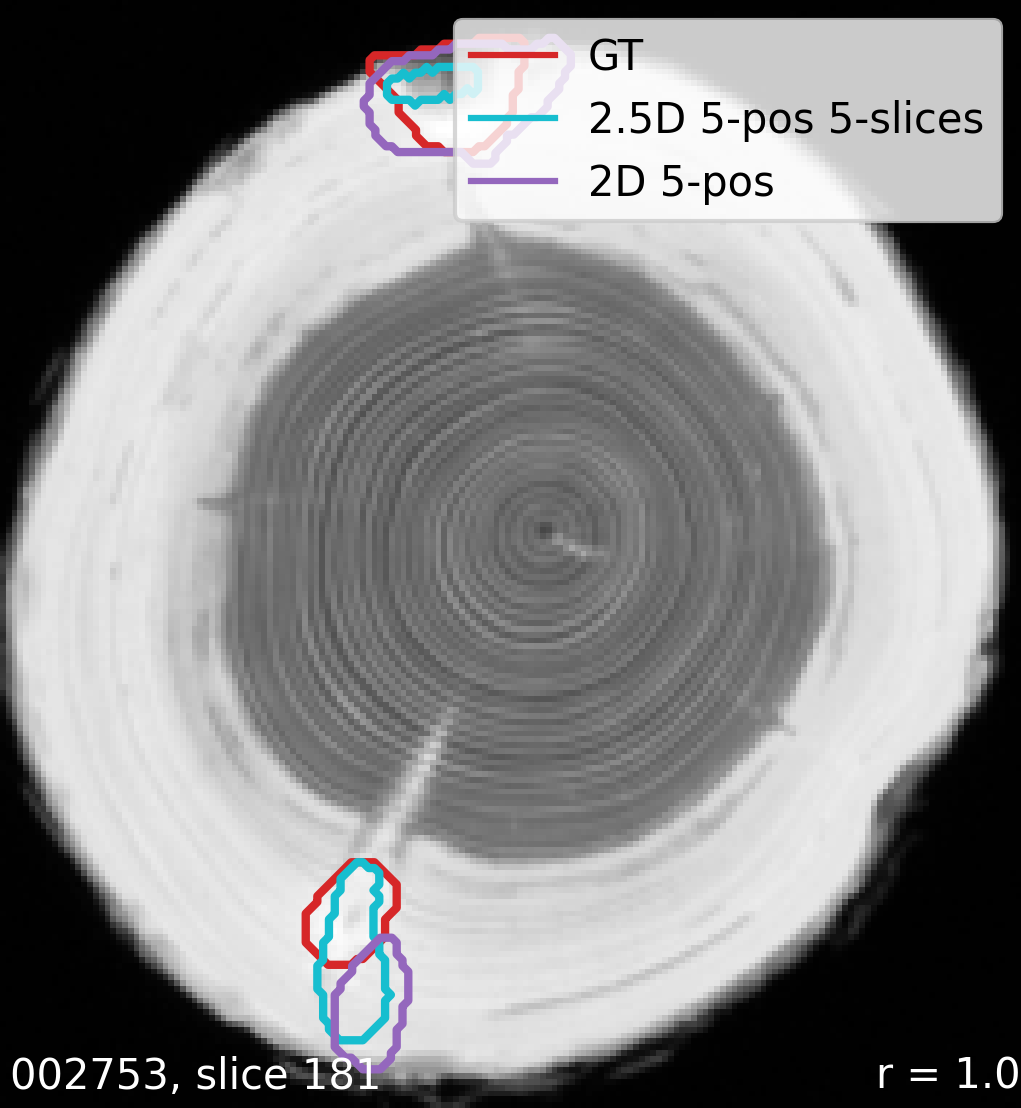} 
\end{tabular}
\caption{Segmentation contours of a knot group in the test sample 002753 by humans (GT), the CT trained U-Net (CT) and the four \ac{LPD} trained U-Nets (2.5D 9-pos 3-slices, 2D 9-pos, 2.5D 5-pos 5-slices, and 2D 5-pos respectively).}
\label{fig:U-Net-labels}
\end{figure}

Additionally, for each knot group in the logs of the test set, the slice-wise Dice scores were extracted, and to account for the varying number of slices in the knot groups, the axial distance along the knot group was normalised and the scores were interpolated.
Figures~\ref{fig:diceprofiles} and \ref{fig:diceprofiles_FBP} show the mean and the standard deviation of these interpolated scores on a normalised knot group distance.
In addition, Figure~\ref{fig:U-Net-labels} shows a sequence of slices through a demonstrative knot group (with normalised distance $r$) in a log of the test set, using the full CT images as a background and the human ground truth (GT) label and inferred label contours superimposed.

Figures~\ref{fig:diceprofiles}a and \ref{fig:diceprofiles_FBP}a show that a greater number of source positions yields a higher Dice score for all sections along knot groups, which was expected due to the increasing amount of information. 
All U-Nets performed well around the middle of a knot group (rows 2-4 Figure~\ref{fig:U-Net-labels}) where knot cross-sections are sufficiently large and where a large share of the knot tissue remains within the heartwood of the log. In the heartwood, contrast is greater than in sapwood due to lower water content and therefore better results were expected.
Towards the origin and end of a knot group, the scores drop sharply.
The U-Nets trained on full CT data and on the 2D \ac{LPD} appear to underestimate knot regions close to their origins, i.e near the pith (first row Figure~\ref{fig:U-Net-labels}).
The U-Nets trained on 2.5D~\ac{LPD} detected at least the presence of knots in these slices, which can be attributed to the 2.5D~\ac{LPD} accounting for neighbouring slices. Figure~\ref{fig:diceprofiles}a shows that this advantage of 2.5D~\ac{LPD} over 2D~\ac{LPD} is more pronounced for variants with 5 source positions than for those with 9 positions.
The end of knot groups are located in the water-rich sapwood (last row Figure~\ref{fig:U-Net-labels}) where the lacking contrast makes a distinction of knots from the background difficult.

Figures~\ref{fig:diceprofiles}b ans \ref{fig:diceprofiles_FBP}b show that the standard deviation in Dice scores increases towards the origin and end of knot groups, which reveals greater inconsistency and uncertainty in the inferred labels in these regions.
Apart from the increased difficulty of segmentation, the uncertainty may be related to inconsistent human labelling of the ground truth in these regions. The labelling was conducted by different individuals with different educational backgrounds and thus their labelling in these difficult regions may have yielded insufficient consistency for generalisation in the models.

Notably, the U-Nets trained on \ac{LPD} variants based on 9 source positions matched and outperformed the U-Nets trained on full CT data close to the origin and end of knot groups.
The models trained on reconstructions with greater levels of detail may have been trying to fit to the ``noisy'' ground truth labels in these images, while the models trained on reconstructions that can resolve fewer details were less prone to overfitting and therefore reached higher Dice scores in these regions.

\section{Conclusion}\label{sec:conclusion}

In this paper, a new 2.5D~\ac{LPD} reconstruction method for tomographic data has been proposed. It bridges the gap between the original \ac{2D} \cite{adler2018learned} and extended \ac{3D} \ac{LPD} methods \cite{rudzusika20223d} suited for \ac{2D} and \ac{3D} geometries. The proposed method is tailored for the case when the scanning geometry is \ac{2D} and for images of an elongated \ac{3D} object with only gradual changes of its internal features along the third dimension.
We evaluated the 2.5D~\ac{LPD} method on wooden logs, which contain biological features, such as sapwood, heartwood, knots, and growth rings, which only change slightly between adjacent cross-sections along the length axis.
The quality of the reconstructions was evaluated by measuring \ac{PSNR} for a varying number X-ray source positions and by performing U-Net-based knot segmentation on the obtained reconstructions. In both types of evaluation, the new method was compared with the original \ac{2D} \ac{LPD} method, which does not take into account the similarity of neighbouring slices while reconstructing the slice of interest.

In terms of \ac{PSNR}, the 2.5D~\ac{LPD} performed slightly better than the original 2D \ac{LPD} for all tested numbers of X-ray source positions, which could be confirmed by visual inspection of the reconstructed images. Even with only 5 source positions, the 2.5D~\ac{LPD} provided useful images, in the sense that all relevant features except for the growth rings were reliably reconstructed. Further examining two different strategies of taking into account several subsequent and preceding slices at a time (``middle'' strategy) instead of only preceding slices (``last'' strategy), showed that the ``middle'' strategy is the most beneficial for a low number of source positions, where it compensates for the lacking richness in data.

For the segmentation of knots, which is the most relevant biological feature in this case, off-the-shelf U-Nets were trained on full CT data and 2D U-Net post-processed \ac{FBP} reconstructions as a benchmark, and on \ac{LPD}-reconstructed data. Overall, the Dice score of the \ac{LPD}-based segmentations was within 15\% of the CT-based segmentation benchmark. All trained U-Nets performed almost equally well in the middle of knot groups, but worse towards the origin and end of knot groups. Due to its ability to take into account several slices at a time, U-Nets based on the 2.5D~\ac{LPD} can perform better than those based on 2D \ac{LPD} in segmenting the beginning of knots groups, in particular for fewer source positions. This circumstance is specifically interesting regarding industrial applicability, as the origin of knots is located in the region of logs from which high-value products are usually sawn.

The observed segmentation behaviour of the trained U-Nets suggests that the quality of the ground truth labelling requires substantial improvements, and it raises the question whether the start of knots could be modelled in a different way, taking into account knowledge about the natural growth behaviour of knots in trees. The former is currently addressed by the creation of a comprehensive dataset with logs CT-scanned in both the wet and dry state, which provides much better contrast in the sapwood region of a log, while the latter is planned to be explored in coming iterations of the presented algorithms.

The present results are based on synthetic projection data derived from laboratory CT images and should therefore be seen as a more idealised case than what can be expected from industrial settings. Nevertheless, they serve as an important proof of principle for the proposed method.
From an industrial perspective, the performance of the 2.5D~\ac{LPD} method in particular is promising, because the ability to obtain 3D volume information about the interior features of a tree with cost-efficient discrete X-ray equipment could enable increased value extraction to a greater number of existing sawmills. Today, this is only possible for sawmills with full CT scanning equipment. To make the presented findings even more interesting for the wood industry, a further reduction in required X-ray source positions may be required.

Future studies should investigate the effects of additionally rotating the log during X-ray image acquisition, which should increase the amount of information, while at the same time reducing the number of X-ray source positions.
In addition, joint reconstruction and segmentation of logs should be investigated \cite{adler2022task} and compared to the sequential approach presented herein.

\appendix

\section{\ac{LPD} implementation details}\label{app:LPD_implementation_details}

The \ac{LPD} methods (2D and 2.5D) were implemented in Python using PyTorch \cite{paszke2017automatic} for neural network layers and training. 
The image and projection spaces were implemented with \ac{ODL} \cite{adler_odl} using ASTRA \cite{van2016fast} as a back-end for evaluating the ray transform and its adjoint. 
The PyTorch implementation of the original \ac{LPD} network which is used in the comparative study of the present paper has exactly the same structure and hyper-parameters as in the original work \cite{adler2018learned}. 
In particular, we used $\NumUnrollingIterates=10$ unrolled iterations and $\Dim=5$ memory channels for both primal and dual variables. 

For the newly proposed 2.5D~\ac{LPD}, we also used $\NumUnrollingIterates=10$ unrolled iterations, while the number of memory channels for both primal and dual variables depended on the number $\totcross$ of consecutive \ac{2D} cross-sections one seeks to account for, which varied between $2$ and $9$, c.f. section \ref{subsec:PSRN_res}.
For both the original \ac{LPD} and the new 2.5D~\ac{LPD}, a three-layer \ac{CNN} was applied in both the image and projection domain at each unrolled iteration, i.e. $\PrimalMappingOriginal_{\param_k^p}$ and $\DualMappingOriginal_{\param_k^d}$ had two hidden convolutional layers with $32$ filters and one convolutional output layer, all using a \ac{2D} kernel size of $7\times7$. The activation functions after the hidden layers were implemented as Parametric Rectified Linear Units (PReLU).
Training was performed for $10^5$ iterations using an initial learning rate set to $10^{-5}$ with cosine annealing decay. 
All experiments were executed on a single NVIDIA RTX A6000 GPU with 49\,140~MiB available memory.

Note that an alternative approach that is not explored in this paper is to to apply \ac{CNN}s with \ac{3D} kernels on sets of neighbouring 2D slices in the image domain, whereas \ac{CNN}s in the sinogram domain are left as in \ac{2D} \ac{LPD}. This is straightforward to implement in the setting with isotropic image voxels and GPU memory requirements for training the resulting \ac{LPD} with \ac{3D} image convolutions will be manageable.

\section{U-Net implementation details}\label{app:U-Net_implementation_details}

The implemented MONAI U-Net \cite{monai2024} architecture features encode-decode paths with skip connections, where strided convolutions in the encode path and strided transpose convolutions in the decode path occur at the start of each block, in contrast to typical U-Net implementations where these down- and upsampling operations occur after each block. Padding and strides are optimised to ensure even division or multiplication of output sizes relative to input dimensions. 
We employed a three-dimensional U-Net architecture, operating on images padded to a size of \(256 \times 256 \times 512\) pixels. 
The model was configured with a single input channel and two output channels, for segmentation of a single target label representing knots, along with the background. 
Structurally, the network comprised five levels of encoding and decoding, featuring channel depths set to [16, 32, 64, 128, 256], and uniform strides of 2 at each level. 
The architecture was enhanced with two residual units, and batch normalisation was applied for feature normalisation.

Training was performed in batches of two under 800 epochs and using random rotations around the log axis and random flipping of the cross-sectional image axes, all with a probability of 0.2. A combined loss function was used, equally weighting Dice and Cross Entropy loss and not counting the Dice score of the background label. The final U-Net model weights were only saved if scoring on the independent validation set (5 samples) improved after a training epoch.

\section{Additional visualisations}\label{app:visuals}

\begin{figure}[ht!]
\centering
\setlength\tabcolsep{0.5pt}
\begin{tabular}{c}
\includegraphics[width=\columnwidth]{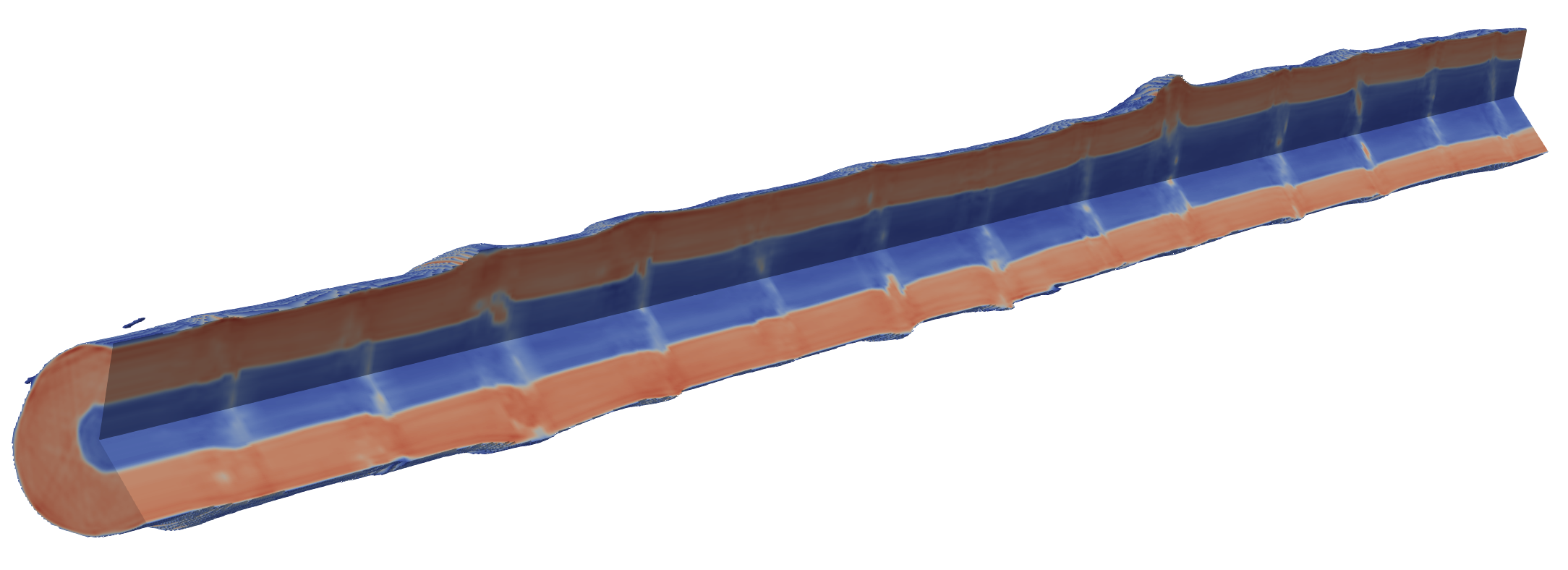}
\\ (a) \\
\includegraphics[width=\columnwidth]{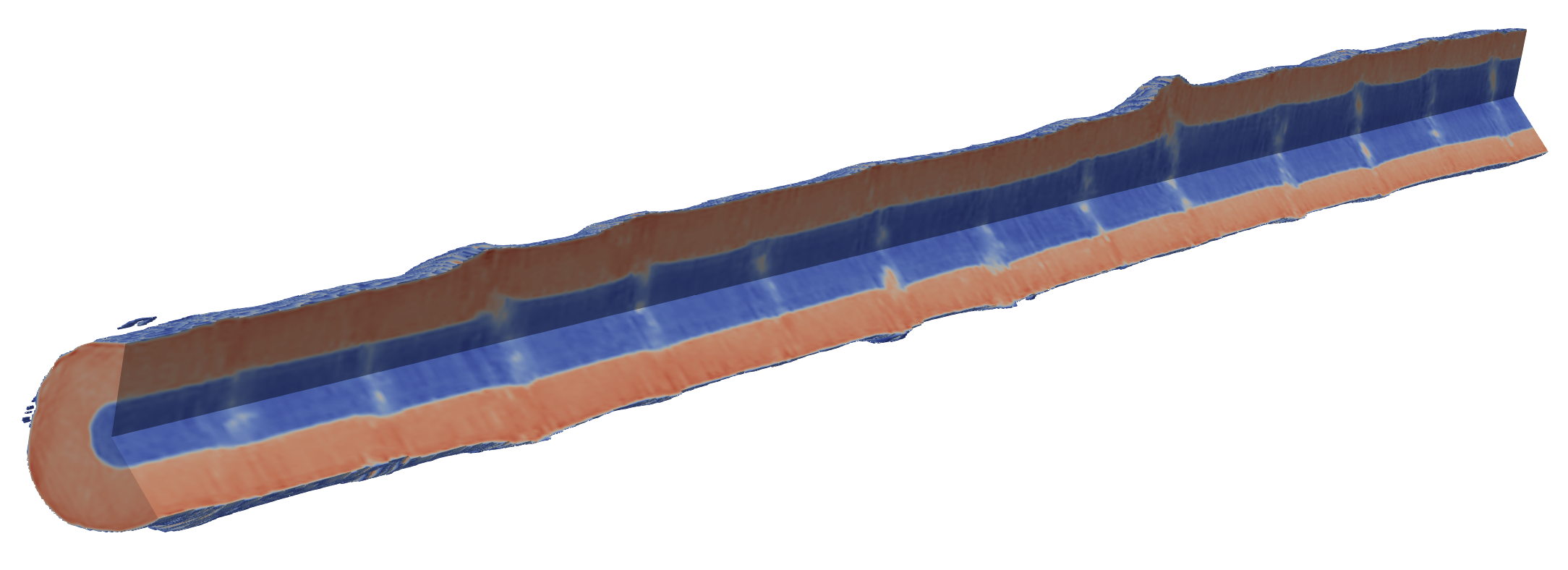}
\\ (b) \\
\includegraphics[width=\columnwidth]{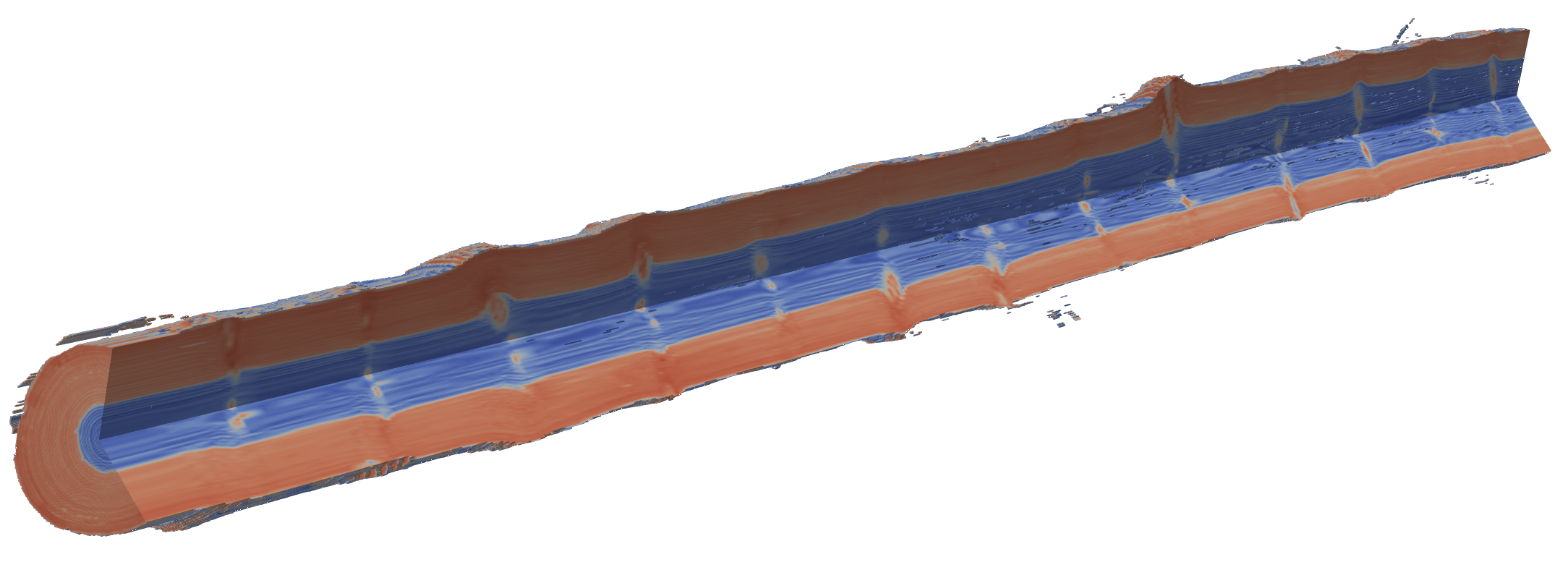}
\\ (c)
\end{tabular}
\caption{3D visualisations of log $001054$, reconstructed with (a) the 2D LPD method with 5 source positions, (b) the 2.5D LPD method with 5 source positions and 5 consecutive slices employing the ``middle'' strategy, and (c) the full CT original.}
\label{fig:3D-vis}
\end{figure}

Figure \ref{fig:3D-vis} compares the 3D appearance of a 2D and 2.5D LPD reconstruction including the original CT reconstruction of the same log.

\section*{Acknowledgment}

The authors thank for the support from the CT WOOD research programme in Skellefteå, Sweden, which is generously supported by the Swedish sawmilling industry, Skellefteå municipality and Luleå University of Technology. The research is partially funded by the Swedish Energy Agency, Vinnova and Formas via the strategic innovation programme RE:Source.

\bibliographystyle{elsarticle-num} 
\bibliography{references}

\begin{thebibliography}{10}
\expandafter\ifx\csname url\endcsname\relax
  \def\url#1{\texttt{#1}}\fi
\expandafter\ifx\csname urlprefix\endcsname\relax\def\urlprefix{URL }\fi
\expandafter\ifx\csname href\endcsname\relax
  \def\href#1#2{#2} \def\path#1{#1}\fi

\bibitem{de2014industrial}
L.~De~Chiffre, S.~Carmignato, J.-P. Kruth, R.~Schmitt, A.~Weckenmann, Industrial applications of computed tomography, CIRP annals 63~(2) (2014) 655--677.

\bibitem{hakkarainen2019undersampled}
J.~Hakkarainen, Z.~Purisha, A.~Solonen, S.~Siltanen, Undersampled dynamic {X}-ray tomography with dimension reduction {Kalman} filter, IEEE Transactions on Computational Imaging 5~(3) (2019) 492--501.

\bibitem{rudin1992nonlinear}
L.~I. Rudin, S.~Osher, E.~Fatemi, Nonlinear total variation based noise removal algorithms, Physica D: nonlinear phenomena 60~(1-4) (1992) 259--268.

\bibitem{donoho2006compressed}
D.~L. Donoho, Compressed sensing, IEEE Transactions on information theory 52~(4) (2006) 1289--1306.

\bibitem{daubechies2004iterative}
I.~Daubechies, M.~Defrise, C.~De~Mol, An iterative thresholding algorithm for linear inverse problems with a sparsity constraint, Communications on Pure and Applied Mathematics: A Journal Issued by the Courant Institute of Mathematical Sciences 57~(11) (2004) 1413--1457.

\bibitem{jin2017deep}
K.~H. Jin, M.~T. McCann, E.~Froustey, M.~Unser, Deep convolutional neural network for inverse problems in imaging, IEEE Transactions on Image Processing 26~(9) (2017) 4509--4522.

\bibitem{monga2021algorithm}
V.~Monga, Y.~Li, Y.~C. Eldar, Algorithm unrolling: {I}nterpretable, efficient deep learning for signal and image processing, IEEE Signal Processing Magazine 38~(2) (2021) 18--44.

\bibitem{arridge2019solving}
S.~Arridge, P.~Maass, O.~{\"O}ktem, C.-B. Sch{\"o}nlieb, Solving inverse problems using data-driven models, Acta Numerica 28 (2019) 1--174.

\bibitem{adler2018learned}
J.~Adler, O.~{\"O}ktem, Learned primal-dual reconstruction, IEEE Transactions on Medical Imaging 37~(6) (2018) 1322--1332.

\bibitem{rudzusika2021invertible}
J.~Rudzusika, B.~Baji{\'c}, O.~{\"O}ktem, C.-B. Sch{\"o}nlieb, C.~Etmann, Invertible {L}earned {P}rimal-{D}ual, in: NeurIPS 2021 Workshop on Deep Learning and Inverse Problems, 2021.

\bibitem{rudzusika20223d}
J.~Rudzusika, B.~Baji{\'c}, T.~Koehler, O.~{\"O}ktem, {3D} helical {CT} {R}econstruction with a {M}emory {E}fficient {L}earned {P}rimal-{D}ual {A}rchitecture, arXiv e-prints (2022) arXiv--2205.

\bibitem{tang2021stochastic}
J.~Tang, S.~Mukherjee, C.-B. Sch{\"o}nlieb, Stochastic primal-dual deep unrolling, arXiv preprint arXiv:2110.10093 (2021).

\bibitem{hammernik2018learning}
K.~Hammernik, T.~Klatzer, E.~Kobler, M.~P. Recht, D.~K. Sodickson, T.~Pock, F.~Knoll, Learning a variational network for reconstruction of accelerated {MRI} data, Magnetic resonance in medicine 79~(6) (2018) 3055--3071.

\bibitem{webber2024diffusion}
G.~Webber, A.~J. Reader, Diffusion models for medical image reconstruction, BJR| Artificial Intelligence 1~(1) (2024) ubae013.

\bibitem{springer2022reconstruction}
S.~Springer, A.~Glielmo, A.~Senchukova, T.~Kauppi, J.~Suuronen, L.~Roininen, H.~Haario, A.~Hauptmann, Reconstruction and segmentation from sparse sequential {X}-ray measurements of wood logs, arXiv preprint arXiv:2206.09595 (2022).

\bibitem{Gergel:2019aa}
T.~Gerge{\v l}, T.~Bucha, M.~Gejdo{\v s}, Z.~Vyhn{\'a}likov{\'a}, Computed tomography log scanning -- high technology for forestry and forest based industry, Central European Forestry Journal 65 (2019) 51--59.
\newblock \href {https://doi.org/10.2478/forj-2019-0003} {\path{doi:10.2478/forj-2019-0003}}.

\bibitem{Rais2017}
A.~Rais, E.~Ursella, E.~Vicario, F.~Giudiceandrea, The use of the first industrial {X}-ray {CT} scanner increases the lumber recovery value: case study on visually strength-graded {Douglas}-fir timber, Annals of Forest Science 74~(2) (2017) 28.
\newblock \href {https://doi.org/10.1007/s13595-017-0630-5} {\path{doi:10.1007/s13595-017-0630-5}}.

\bibitem{ursella2018}
E.~Ursella, F.~Giudiceandrea, M.~Boschetti, \href{https://www.ndt.net/?id=21911}{A {Fast} and {Continuous} {CT} scanner for the optimization of logs in a sawmill}, in: 8th {Conference} on {Industrial} {Computed} {Tomography} ({iCT}) 2018, Vol.~23, e-Journal of Nondestructive Testing, Wels, Austria, 2018.
\newline\urlprefix\url{https://www.ndt.net/?id=21911}

\bibitem{LONGUETAUD2012}
F.~Longuetaud, F.~Mothe, B.~Kerautret, A.~Krähenbühl, L.~Hory, J.~Leban, I.~Debled-Rennesson, Automatic knot detection and measurements from x-ray ct images of wood: A review and validation of an improved algorithm on softwood samples, Computers and Electronics in Agriculture 85 (2012) 77--89.
\newblock \href {https://doi.org/https://doi.org/10.1016/j.compag.2012.03.013} {\path{doi:https://doi.org/10.1016/j.compag.2012.03.013}}.

\bibitem{JOHANSSON2013}
E.~Johansson, D.~Johansson, J.~Skog, M.~Fredriksson, Automated knot detection for high speed computed tomography on pinus sylvestris l. and picea abies (l.) karst. using ellipse fitting in concentric surfaces, Computers and Electronics in Agriculture 96 (2013) 238--245.
\newblock \href {https://doi.org/https://doi.org/10.1016/j.compag.2013.06.003} {\path{doi:https://doi.org/10.1016/j.compag.2013.06.003}}.

\bibitem{Seplveda2002}
P.~Sepúlveda, J.~Oja, A.~Gr\"{o}nlund, Predicting spiral grain by computed tomography of norway spruce, Journal of Wood Science 48~(6) (2002) 479–483.
\newblock \href {https://doi.org/http://doi.org/10.1007/bf00766643} {\path{doi:http://doi.org/10.1007/bf00766643}}.

\bibitem{Ekevad2004}
M.~Ekevad, Method to compute fiber directions in wood from computed tomography images, Journal of Wood Science 50~(1) (2004) 41–46.
\newblock \href {https://doi.org/http://doi.org/10.1007/s10086-003-0524-z} {\path{doi:http://doi.org/10.1007/s10086-003-0524-z}}.

\bibitem{Hansson2016}
L.~Hansson, J.~Couceiro, B.-A. Fjellner, Estimation of shrinkage coefficients in radial and tangential directions from ct images, Wood Material Science \& Engineering 12~(4) (2016) 251–256.
\newblock \href {https://doi.org/http://doi.org/10.1080/17480272.2016.1249405} {\path{doi:http://doi.org/10.1080/17480272.2016.1249405}}.

\bibitem{Huber2022}
J.~A.~J. Huber, O.~Broman, M.~Ekevad, J.~Oja, L.~Hansson, A method for generating finite element models of wood boards from x-ray computed tomography scans, Computers \& Structures 260 (2022) 106702.
\newblock \href {https://doi.org/https://doi.org/10.1016/j.compstruc.2021.106702} {\path{doi:https://doi.org/10.1016/j.compstruc.2021.106702}}.

\bibitem{chambolle2011first}
A.~Chambolle, T.~Pock, A {F}irst-{O}rder {P}rimal-{D}ual {A}lgorithm for {C}onvex {P}roblems with {A}pplications to {I}maging, Journal of mathematical imaging and vision 40~(1) (2011) 120--145.

\bibitem{gronlund1995}
S.~Grundberg, A.~Gr{\"o}nlund, U.~Gr{\"o}nlund, The {S}wedish {S}tem {B}ank – {A} database for different silvicultural and wood properties, Tech. rep., Luleå University of Technology, Skellefteå, Sweden (1995).

\bibitem{monai2024}
{The MONAI Consortium}, \href{https://doi.org/10.5281/zenodo.4323059}{Project {MONAI}} (Dec. 2020).
\newblock \href {https://doi.org/10.5281/zenodo.4323059} {\path{doi:10.5281/zenodo.4323059}}.
\newline\urlprefix\url{https://doi.org/10.5281/zenodo.4323059}

\bibitem{adler_odl}
J.~Adler, H.~Kohr, O.~{\"O}ktem, Operator discretization library ({ODL}) (2017).

\bibitem{wang2004image}
Z.~Wang, A.~C. Bovik, H.~R. Sheikh, E.~P. Simoncelli, Image quality assessment: from error visibility to structural similarity, IEEE transactions on image processing 13~(4) (2004) 600--612.

\bibitem{residualUnet2019}
E.~Kerfoot, J.~Clough, I.~Oksuz, J.~Lee, A.~P. King, J.~A. Schnabel, Left-{V}entricle {Q}uantification {U}sing {R}esidual {U}-{N}et, in: M.~Pop, M.~Sermesant, J.~Zhao, S.~Li, K.~McLeod, A.~Young, K.~Rhode, T.~Mansi (Eds.), Statistical Atlases and Computational Models of the Heart. Atrial Segmentation and {LV} Quantification Challenges, Springer International Publishing, Cham, 2019, pp. 371--380.

\bibitem{adler2022task}
J.~Adler, S.~Lunz, O.~Verdier, C.-B. Sch{\"o}nlieb, O.~{\"O}ktem, Task adapted reconstruction for inverse problems, Inverse Problems 38~(7) (2022) 075006.

\bibitem{paszke2017automatic}
A.~Paszke, S.~Gross, S.~Chintala, G.~Chanan, E.~Yang, Z.~DeVito, Z.~Lin, A.~Desmaison, L.~Antiga, A.~Lerer, Automatic differentiation in {P}y{T}orch, in: 31st Conference on Neural Information Processing Systems (NIPS 2017), Long Beach, CA, USA, 2017.

\bibitem{van2016fast}
W.~Van~Aarle, W.~J. Palenstijn, J.~Cant, E.~Janssens, F.~Bleichrodt, A.~Dabravolski, J.~De~Beenhouwer, K.~J. Batenburg, J.~Sijbers, Fast and flexible {X}-ray tomography using the {ASTRA} toolbox, Optics express 24~(22) (2016) 25129--25147.

\end{thebibliography}

\end{document}